\documentclass[runningheads]{llncs}

\usepackage{eccv}

\usepackage{eccvabbrv}

\usepackage{graphicx}
\usepackage{booktabs}
\usepackage{array}
\usepackage{bbm}
\usepackage{arydshln}
\usepackage{multirow}
\usepackage{adjustbox}
\usepackage{makecell}
\usepackage{pifont}
\usepackage{textcomp} 
\usepackage{amsmath}

\newcommand{\cmark}{\ding{51}} 
\newcommand{\xmark}{\ding{55}} 
\usepackage{makecell}

\usepackage{tcolorbox}
\tcbuselibrary{listings,breakable,skins}
\tcbset{
  mylisting/.style={
    breakable,
    enhanced,
    colback=black!3,
    colframe=black!60,
    sharp corners,
    listing only,
    listing options={
      basicstyle=\ttfamily\footnotesize,
      columns=fullflexible,
      keepspaces=true,
      breaklines=true,
      breakindent=0pt,
      breakatwhitespace=true,
      literate=
        {≈}{{$\approx$}}1
        {°}{{\textdegree}}1
    }
  }
}

\usepackage[accsupp]{axessibility}  

\usepackage{hyperref}

\usepackage{orcidlink}

\begin{document}

\title{360CityArena: A Realistic Virtual Urban Navigation Benchmark for Embodied Agents} 

\titlerunning{360CityArena}

\author{Kenta Watanabe\and
Atsuyuki Miyai \and
Mizuki Takenawa \and\\
Kiyoharu Aizawa \and
Toshihiko Yamasaki
}

\authorrunning{K.~Watanabe et al.}

\institute{ The University of Tokyo, Tokyo, Japan\\ \email{\{k\_watanabe,miyai,yamasaki\}@cvm.t.u-tokyo.ac.jp}\\ \email{\{takenawa,aizawa\}@hal.t.u-tokyo.ac.jp}\\
\url{https://360mm-team.github.io/360CityArena/}
}

\maketitle

\begin{figure*}[h]
\vspace{-20pt}
\centering
    \includegraphics[width=0.99\linewidth]{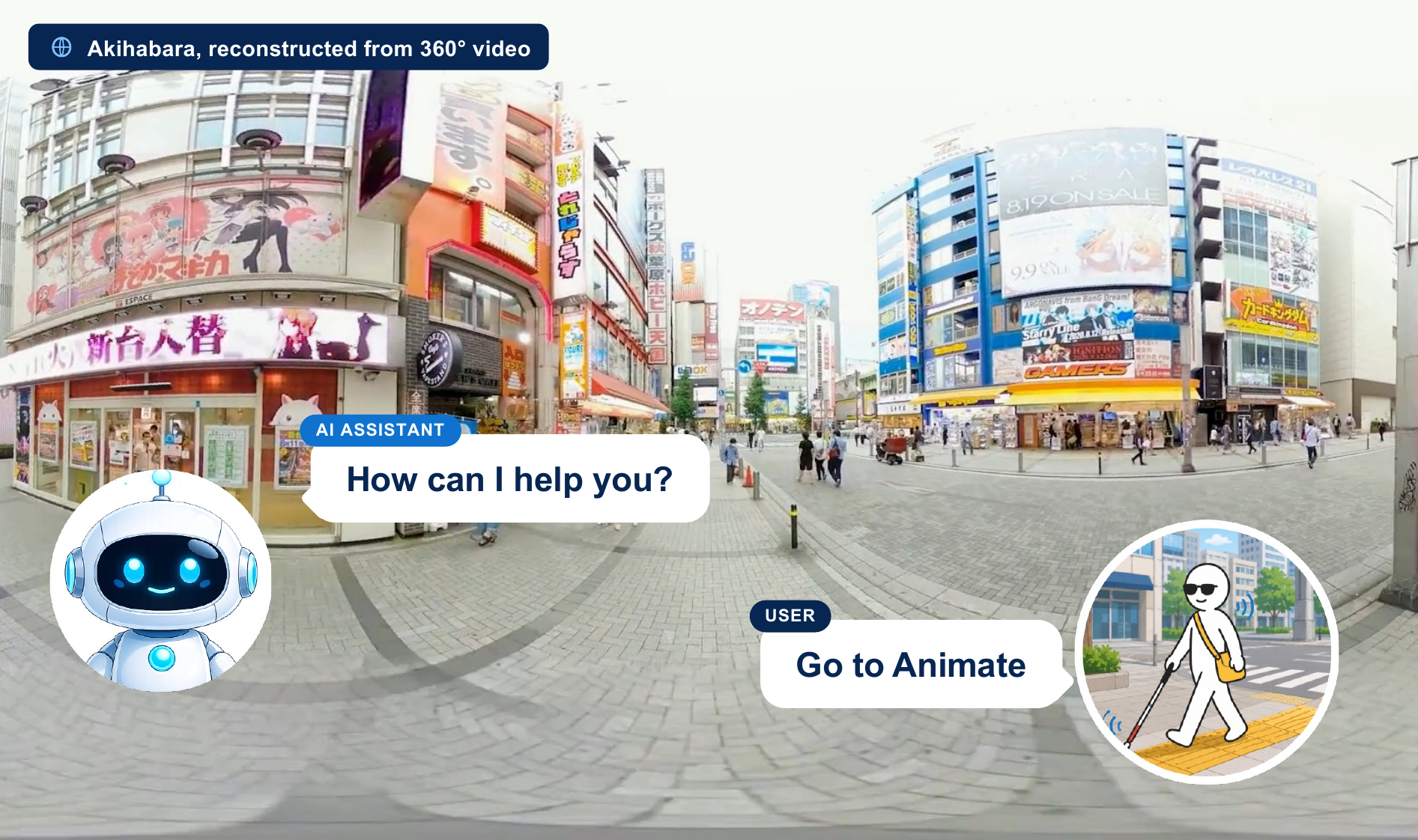}\\
    \vspace{-5pt}
    \caption{\small \textbf{360CityArena}. We introduce a benchmark for evaluating embodied agents in a photorealistic reconstruction of Akihabara, Tokyo, Japan, built from interconnected 360\textdegree~video trajectories. The benchmark covers realistic urban streets and evaluates agents on diverse tasks requiring environment understanding, path reasoning, and spatial reasoning.}
    \vspace{-5pt}
    \label{fig:fig_teaser}
\end{figure*}

\vspace{-20pt}

\begin{abstract}
We present 360CityArena, a benchmark for evaluating the urban exploration capabilities of embodied agents within a photorealistic environment constructed from 360\textdegree~videos. 
Existing outdoor benchmarks either lack sufficient photorealism or complexity, resulting in a considerable gap from real-world urban environments.
360CityArena is built on a realistic reconstruction of the Akihabara district in Tokyo, Japan, using 602~360\textdegree~video segments covering 85 streets, and consists of 175 meticulously human-crafted tasks. 
It encompasses three task categories: Environment Understanding, Path Reasoning, and Spatial Reasoning, covering fundamental abilities required for urban exploration, such as localization, landmark search, path planning, and relational spatial reasoning, thereby enabling comprehensive evaluation in realistic urban scenes.
Our evaluation using state-of-the-art LMM-based agents shows that even the strongest model, Gemini 2.5 Flash, performs far below human level (human: 77.3\% vs. Gemini 2.5 Flash: 17.1\%), revealing substantial challenges that remain in city-scale embodied navigation and reasoning. 
360CityArena provides a necessary and challenging testbed for photorealistic urban-district navigation and spatial reasoning.

\end{abstract}

\section{Introduction}
\label{sec:intro}
Imagine a future where an AI assistant is a natural part of city life. It can help you when you get lost, guide blind people to where they need to go, and provide support whenever someone needs it. This kind of assistant would make cities more accessible, reduce the difficulties of traveling, and help create a more inclusive society. One promising approach toward achieving this vision is Embodied AI, in which agents perform complex tasks based on perception of their surrounding environment~\cite{das2018eqa,anderson2018evaluation, duan2022survey, yang2025survey}.
Realizing this vision will require major advances in evaluating and testing such systems across a wide range of tasks within realistic urban environments that capture the visual complexity of real streets.

A central problem in city exploration research is the lack of benchmarks that evaluate agents in environments that realistically reflect wide-area navigation in real urban spaces. Existing 3D simulators fall short in reconstructing city scenes with sufficient photorealism and complexity~\cite{ji2025towards, gao2024embodiedcity, wu25metaurban,dosovitskiy2017carla}.
Although environments built from Google Street View provide realistic imagery~\cite{mirowski2018learning, Chen_2019_CVPR, yang25virl, feng25citybench}, they lack dynamic elements and do not offer continuous, fully navigable spaces.
To address these issues, recent studies have explored converting real-world urban videos into simulation environments, enabling more realistic and interactive experiences~\cite{xie2025vid2sim}. However, the video clips used in this method are short, making them unsuitable for exploring extended real street networks.
In parallel, recent work has studied navigation from different modalities and viewpoints, such as aerial VLN over cities and map-only evaluation of route planning~\cite{lee2025citynav, paz-argaman19run, xing2025map}.
These lines of work are complementary, but they still do not capture ground-level, egocentric city exploration in realistic street scenes with rich dynamics.
To further advance the field, we need a benchmark that covers tasks within highly realistic and dynamic urban street environments, while supporting smooth egocentric navigation.
\begin{figure*}[t]
\centering
    \includegraphics[width=0.99\linewidth]{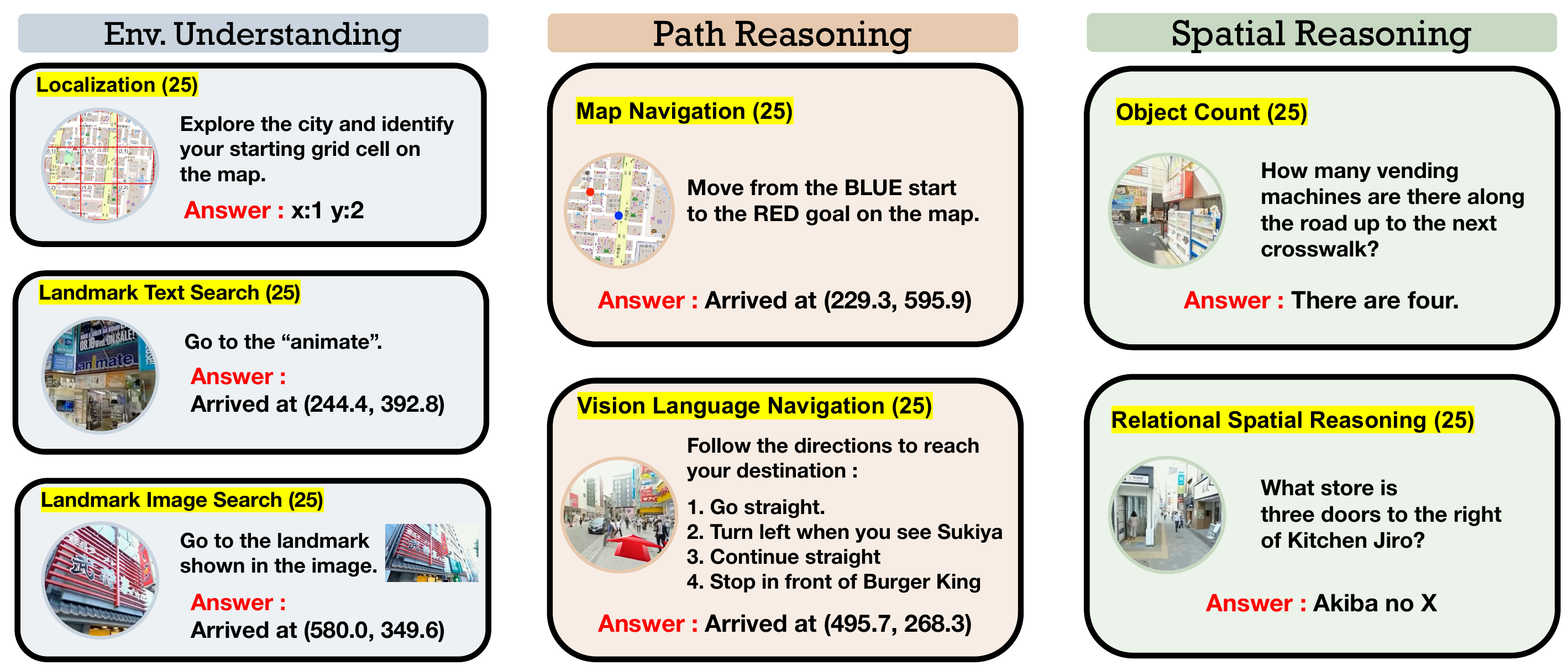}
   \caption{\textbf{Examples in each task type in 360CityArena.} (i) Environment Understanding tasks include \textit{Localization}, \textit{Landmark Search with Language}, and \textit{Landmark Search with Image}, where the agent infers its location or navigates to a specified landmark. (ii) Path Reasoning tasks evaluate the agent's ability to plan and execute routes, such as following map-based paths or vision-language navigation. (iii) Spatial Reasoning tasks assess relational understanding and quantitative perception, including identifying spatial relations between landmarks and counting objects in the environment.}
    \label{fig:task_type}
\end{figure*}

In this paper, we propose \textbf{360CityArena}, a benchmark for assessing embodied agents in a photorealistic real-world urban district represented as an interconnected pose graph of 360\textdegree~video trajectories.
360CityArena is built upon a Realistic Virtual World environment~\cite{takenawa2025building} that recreates Akihabara, a major urban district in Tokyo, and includes 175 manually created tasks. 
The environment spans approximately 750 meters north–south and 650 meters east–west, covering 85 interconnected streets.
Each task is grounded in real city structures and landmarks, encouraging agents to perform practical, exploration-based activities.

As shown in \Cref{fig:task_type}, 360CityArena consists of three task types: (i) \textit{Environment Understanding}, (ii) \textit{Path Reasoning}, and (iii) \textit{Spatial Reasoning}.
(i) \textit{Environment Understanding} targets visual recognition and self-localization, measuring an agent's perceptual understanding of the environment.
(ii) \textit{Path Reasoning} focuses on path planning and decision-making ability within the real spatial layout of the environment.
(iii) \textit{Spatial Reasoning} evaluates geospatial reasoning through navigation.
At a finer level, the benchmark systematically measures seven distinct capabilities. Furthermore, each task is categorized into Easy, Medium, and Hard, making it possible to evaluate agents according to difficulty level. Through these systematically constructed tasks, 360CityArena enables a comprehensive assessment of an Embodied Agent's ability to perceive, reason, and act in real-world urban environments.

Our experiments show that state-of-the-art LMM-based agents remain far below human performance across all tasks, especially on Map Navigation, Object Count, and Relational Spatial Reasoning. This demonstrates both the value of 360CityArena as a comprehensive benchmark for realistic urban settings and the need for stronger city-scale agents.
We also evaluate removing explicit self-location information (\eg~a map). Performance does not degrade consistently and even improves in some tasks, suggesting that the manner of providing location priors can affect exploration strategies and failure modes, informing future model design.

The contributions of our paper are summarized as follows:
\begin{itemize}
      \item \textbf{Proposing 360CityArena}:
      We introduce \textbf{360CityArena}, an urban-district benchmark built on a photorealistic Virtual World reconstruction of Tokyo's Akihabara district from 360\textdegree~videos. The benchmark provides seven task types across three categories: Environment Understanding, Path Reasoning, and Spatial Reasoning, enabling comprehensive evaluation of embodied agents in a realistic urban environment with an interconnected street network. Beyond evaluation, 360CityArena would also serves as a practical guideline for constructing training data in photorealistic virtual worlds for embodied AI.
      \item \textbf{Benchmarking Recent LMMs}:
      We build embodied agents based on several proprietary LMMs and open-source vision-language models, and benchmark them on 360CityArena. The results reveal substantial performance gaps compared to humans across all tasks and clear performance degradation as task difficulty increases.
      \item \textbf{Comprehensive Analysis}: Through extensive qualitative and quantitative analyses, we investigate how input modality (language versus images), the presence of self-location information, and model- and task-specific failure patterns influence agent behavior and performance. These analyses offer insights into the roles of visual cues and self-localization in urban embodied tasks.
\end{itemize}

\section{Related Work}
\label{sec:related_work}

\begin{table*}[t]
\caption{Comparison of urban navigation environments -- realism, structural complexity, dynamics, interactivity, and exploration capability.}
\centering
\small
\setlength{\tabcolsep}{3.5pt}
\renewcommand{\arraystretch}{1.1}
\resizebox{\textwidth}{!}{%
\begin{tabular}{l l c c c c c l}
\toprule
Environment & Category &
\makecell{Photo-\\realism} &
\makecell{Structural\\complexity} &
Dynamics &
Interaction &
\makecell{District-scale\\exploration} &
Motion \\
\midrule
EmbodiedCity~\cite{gao2024embodiedcity}      & 3D simulator & Low  & Low  & Medium & \cmark & \cmark & Continuous \\
MetaUrban~\cite{wu25metaurban}           & 3D simulator & Low  & Medium  & Medium & \cmark & \cmark & Continuous \\
CARLA~\cite{dosovitskiy2017carla}               & 3D simulator & Low  & Low  & Low & \cmark & \cmark & Continuous (clip) \\
Vid2Sim~\cite{xie2025vid2sim}             & Video-to-sim & High & High & Medium & \cmark & \xmark & Continuous \\
StreetLearn~\cite{mirowski2018learning}         & GSV-based    & High & High & Low    & \xmark & \cmark & Discrete \\
\midrule
\textbf{360CityArena (Ours)} & 360\textdegree~video & High & High & High & \xmark & \cmark & \makecell{Continuous (trajectory)} \\
\bottomrule
\end{tabular}}
\label{tab:env_comparison}
\end{table*}

\noindent\textbf{Embodied Agent.}
An Embodied Agent is an AI system that follows language instructions and completes tasks by perceiving the environment through camera views and sensor information. Recent advances in LMMs have greatly improved multimodal integration and reasoning, making them a strong foundation for embodied agents~\cite{ichter2022do, huang2022zeroshot, zitkovich2023rt2, li2024embodiedinterface}. Embodied-agent research spans real robots~\cite{Zaffar2021VPRBench, shah2023vint, kawaharazuka2025vla}, indoor simulators~\cite{savva2019habitat, khanna2024goat, yang2025embodiedbench}, and outdoor simulators~\cite{Chen_2019_CVPR, xie2025vid2sim}. While real robots are most realistic, data collection is costly and often environment-specific, limiting generalization. Simulation environments therefore attract attention for their diversity and reproducibility~\cite{duan2022survey}. Indoor simulators ease data collection but have limited scale and simple layouts~\cite{savva2019habitat}, motivating a shift toward larger, more complex outdoor settings with dynamic elements such as pedestrians and vehicles~\cite{mirowski2018learning,yang25virl}.

\noindent\textbf{Navigation Agents in Outdoor Environments.}
Recent advances in simulation environments have broadened embodied navigation research beyond indoor settings~\cite{savva2017minos, savva2019habitat, majumdar22zson, khanna2024goat, puig2024habitat3, yang2025embodiedbench} to large-scale outdoor environments.
Outdoor navigation tasks are commonly distinguished by how the goal is specified: point-goal navigation, where the agent is given target coordinates~\cite{mirowski2018learning, wu25metaurban, liu2025citywalker}; image-goal navigation, where the target is provided as a reference image~\cite{jiao2025litevloc, ji2025towards}; object-goal navigation, where the goal is described in language (\eg~a landmark)~\cite{brahmbhatt2017deepnav, hong2025embodied}; and Vision-Language Navigation (VLN), where the agent follows natural-language instructions~\cite{Chen_2019_CVPR, li2024vln, yang25virl}.
Some work instead evaluates route planning directly on maps without embodied simulation~\cite{paz-argaman19run, xing2025map, yang25virl}.
In parallel, CityNav~\cite{lee2025citynav} studies real-world navigation from an \emph{aerial} viewpoint, which is complementary but differs from \emph{ground-level} exploration with egocentric street scenes.
Separately, map understanding has been explored outside embodied navigation, including text-based map representations for grounding LLM planning (Tag Map~\cite{zhang2024tagmap}) and VLM benchmarks for advanced map queries (MAPWise~\cite{mukhopadhyay2025mapwise}).

While point-goal, image-goal, and object-goal tasks emphasize efficient target reaching via visual or semantic cues, VLN additionally demands higher-level reasoning to resolve linguistic ambiguity and landmark references.
However, aerial-VLN datasets and map-only evaluations do not jointly capture egocentric navigation and urban reasoning in realistic street environments.
To address this gap, we introduce an urban-district benchmark that unifies these tasks with (i) photorealistic, dynamic street-level observations, (ii) multi-task diagnostics spanning environment understanding, path reasoning, and spatial reasoning, and (iii) controlled comparisons between language- and image-goal landmark search under matched conditions.

\noindent\textbf{Real-world Simulation Environments.}
Prior work has explored 3D scene reconstruction, Google Street View–based environments, and 3D simulators for building real-world simulation settings.
For 3D scene reconstruction, methods built on Neural Radiance Fields~\cite{mildenhall2020nerf} and Gaussian Splatting~\cite{kerbl3Dgaussians} have led to numerous extensions~\cite{barron2022mipnerf360, Yang2023FreeNeRF, xu2024fewshotnerfadaptiverendering, zhang2024gaussian}. Despite these advancements, most approaches remain limited to producing static scenes and do not provide fully interactive environments~\cite{xie2025vid2sim}.
City-scale reconstructions in conventional 3D simulators can procedurally generate diverse variations and support rich agent–environment interactions, but fall short in photorealism and structural complexity~\cite{gao2024embodiedcity, ji2025towards, wu25metaurban, liu2026sidewalkbench}. 
Environments constructed from Google Street View (GSV) enable diverse and highly realistic city-scale representations that are difficult to achieve with traditional 3D simulation~\cite{mirowski2018learning, Chen_2019_CVPR, yang25virl, feng25citybench, wang2026cityseeker}. However, these environments from GSV do not contain dynamic elements, and the inherent discontinuity between panoramas creates a gap from real-world continuous navigation~\cite{mirowski2018learning}.
To address this, a recent effort has attempted to convert real urban videos into interactive simulation environments~\cite{xie2025vid2sim}. While this approach is promising, this environment relies on short video clips, making them unsuitable for large-scale city exploration.
Takenawa \etal~\cite{takenawa2025building} introduced a Realistic Virtual World (RVW) generated from large collections of 360\textdegree~videos, providing a photorealistic and dynamic city-scale environment with real pedestrian and vehicle motions. We leverage this RVW to build a benchmark for realistic, city-scale exploration by embodied agents. While navigation within each filmed trajectory is continuous and smooth, discontinuities arise at trajectory boundaries, and the environment does not support physical interaction because it is constructed from pre-recorded video. We summarize key properties of representative real-world urban environments in Table~\ref{tab:env_comparison}.

\section{The 360CityArena}

\begin{figure}[t]
\centering
    \includegraphics[width=0.99\linewidth]{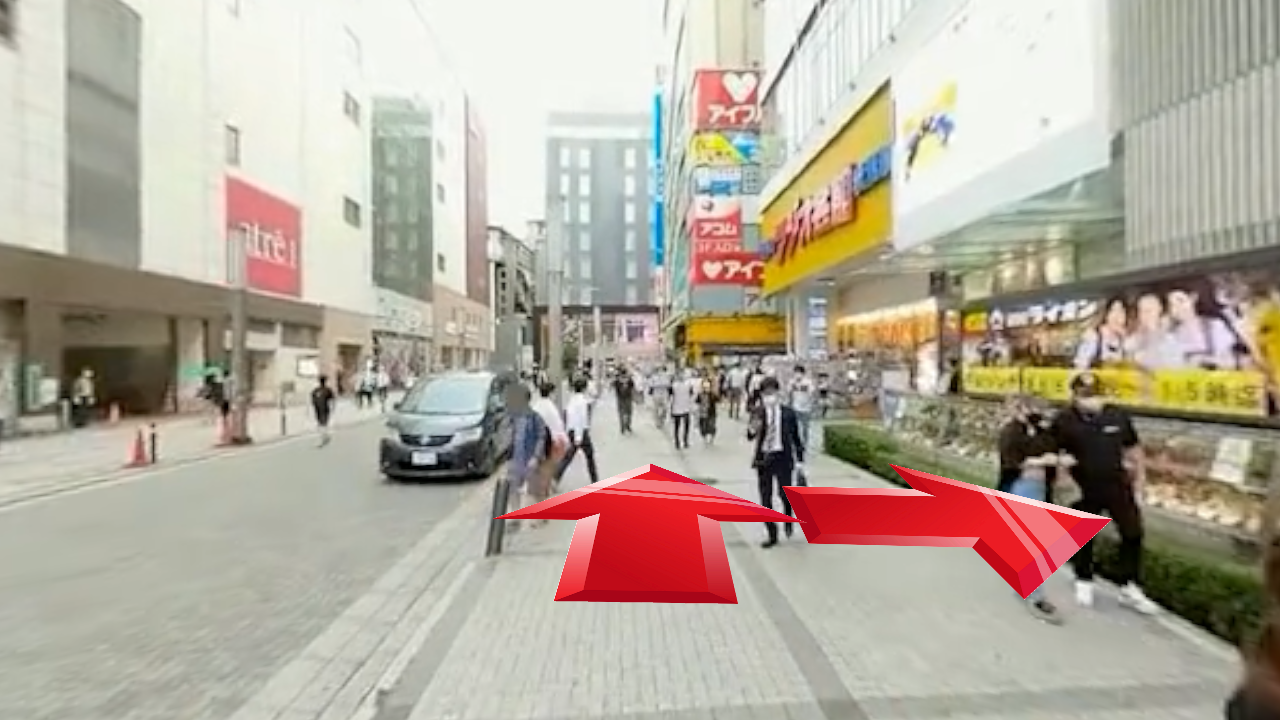}\\
   \caption{\textbf{Example of the visual observations in 360CityArena.} The agent moves through the city while scanning its surroundings. At branching points, it must choose a route.}
    \label{fig:example_visual}
\end{figure}

We introduce 360CityArena, a novel benchmark meticulously designed to evaluate an agent's ability to explore a realistic urban district across a broad range of tasks (see \Cref{fig:example_visual}). The benchmark is built on a Realistic Virtual World reconstructed from the actual Akihabara district in Japan and consists of 175 tasks organized into three major categories and seven subcategories.

We first describe the Akihabara virtual environment in \Cref{subsec:bench_env}. We then provide detailed explanations of each task category in \Cref{subsec:bench_task_category}. The task construction process is outlined in \Cref{subsec:bench_construction}, followed by the evaluation protocol in \Cref{subsec:bench_evaluation}.

\subsection{Akihabara Virtual Environment}
\label{subsec:bench_env}
The environment used in 360CityArena is a Realistic Virtual World of Akihabara constructed from 602 360\textdegree~video segments over 85 streets~\cite{takenawa2025building}. The videos are projected onto a spherical surface and organized into a navigable pose graph in Unity, enabling agents to traverse the captured area. Agents therefore navigate along prerecorded 360\textdegree{} video trajectories represented as a pose graph, rather than moving freely to arbitrary 3D positions or physically interacting with the environment. The resulting pose graph forms a single connected component with 193 nodes and 305 edges, and has a mean branching degree of 3.16.

In terms of geographic coverage, 360CityArena spans approximately 750\,m north–south and 650\,m east–west around Akihabara Station. Akihabara, originally an electronics district, has become a major commercial and cultural hub with dense electronics and anime/game-related retail and tourist attractions. The area features diverse streetscapes with narrow sidewalks, complex building layouts, abundant outdoor ads, and electronic signboards, creating a visually dense, information-rich scene. Each street is captured twice (one per direction), reflecting direction-dependent visual changes. These characteristics make Akihabara well suited for evaluating embodied agents’ perception and navigation in a realistic urban simulation.

Although 360CityArena currently includes only one city, the representation is general. Following the Movie Map paradigm~\cite{sugimoto2020moviemap}, similar city-scale environments can be built by geo-localizing 360\textdegree{} frames with GPS/SLAM and a map, identifying intersections, and connecting snippets into a traversable pose graph. Our city-agnostic graph augmented with 360\textdegree{} observations is directly compatible with such environments.

\subsection{Task Categories}
\label{subsec:bench_task_category}
Our benchmark consists of three major categories and seven subcategories, with examples shown in \Cref{fig:task_type}. Each subcategory contains 25 tasks. We describe each category in detail below.

\noindent\textbf{1. Environment Understanding.}
This category evaluates an agent’s ability to perceive and understand its surroundings, covering visual perception, memory, and language-grounded scene understanding. It includes three subcategories: \textit{Localization}, \textit{Landmark Search with Language} (object-goal), and \textit{Landmark Search with Image} (image-goal).
In \textit{Localization}, the agent infers its initial position on a 5×5 grid from visual observations. Unlike GeoGuessr-style geolocalization, which predicts broad regions from scenes~\cite{Haas_2024_CVPR}, our task operates at a much smaller scale and relies on town layout and local landmarks.
The two Landmark Search tasks require navigating to a specified landmark, provided either in natural language or as an image; aside from the input modality, they share identical settings, enabling controlled comparison of modality effects.

\noindent\textbf{2. Path Reasoning.}
This category evaluates an agent's ability to plan and explore based on spatial understanding. It measures not only the agent's planning and decision-making abilities, but also its ability to execute those plans in accordance with the spatial layout of the environment.
The subcategories in this task are \textit{Map Navigation} and \textit{Vision-Language Navigation} (VLN).
\textit{Map Navigation} requires the agent to choose an optimal route using map information and navigate from a specified start point to a designated goal point.
\textit{Vision-Language Navigation} requires the agent to interpret and follow multi-step natural-language instructions and move to the target accordingly.

\noindent\textbf{3. Spatial Reasoning.}
This category evaluates an agent's ability to understand the structural layout of objects and the positional relationships between landmarks in the environment. It measures core spatial perception and logical reasoning capabilities.
The subcategories in this task are \textit{Relational Spatial Reasoning} and \textit{Object Count}.
\textit{Relational Spatial Reasoning} requires the agent to infer relative spatial relationships between landmarks. Given a reference landmark and a specified relation, the agent must identify the landmark that satisfies that relation.
\textit{Object Count} is a quantity-estimation task in which the agent must accurately count the number of specified objects within a designated area.

\subsection{Benchmark Construction}
\label{subsec:bench_construction}
Our construction involves eight annotators.
We assigned two annotators to each of the seven sub-tasks. To ensure consistency in task quality across different tasks, one of the authors was assigned to all tasks.
To ensure the quality of the benchmark, we selected annotators who had actually visited Akihabara in person. They interacted with the environment directly in Unity and created tasks, and verified that each task was solvable. The overall task construction process required approximately 60 hours.

Each task is labeled Easy, Medium, or Hard by annotators based on distance, instruction ambiguity, landmark visibility, and required exploration, rather than fixed step-count thresholds. To support these labels, task-relevant statistics show monotonic increases from Easy to Hard: Map Navigation path length and decision points are 124/247/347 m and 3.3/6.4/9.0, Object Count ground-truth counts are 3.5/4.6/9.0, and VLN instruction steps are 4.4/5.9/7.4. Borderline cases may still involve subjectivity.

\subsection{Evaluation Protocol}
We define a set of evaluation criteria to assess agent performance. Our benchmark adopts four evaluation protocols, each designed to capture different aspects of task success. We detail each protocol and its corresponding task categories below.

\noindent\textbf{1.~exact\_match.}
\texttt{exact\_match} is applied to tasks whose outputs are explicitly defined as numerical values (grid coordinates) or textual strings~\cite{zhou2023webarena, miyai2025webchorearena}. Under this metric, the agent's final textual output is evaluated, and the prediction is considered correct only when it exactly matches the answer text. In our benchmark, the \textit{Localization} task uses this evaluation metric.

\noindent\textbf{2.~fuzzy\_match.}
\texttt{fuzzy\_match} leverages a language model (GPT-5 in our implementation) to assess whether the output is semantically equivalent to the ground truth~\cite{zhou2023webarena, miyai2025webchorearena}. Importantly, the GPT-5 evaluator is fully isolated from the agent and only compares the final output with the ground-truth answer, preventing any information leakage.
In the \textit{Relational Spatial Reasoning} task, we employed this evaluation metric and additionally performed manual verification. The GPT-5 judge achieved 97.9\% agreement with human majority votes ($\kappa=0.937$), close to the inter-annotator agreement ($\kappa=0.984$), supporting its use for evaluation automation.

\noindent\textbf{3.~coordinate\_match.}
\texttt{coordinate\_match} is used for tasks in which the agent must output its final position after moving through the environment. The coordinates are obtained directly from Unity. This metric is applied to \textit{Map Navigation}, \textit{VLN}, and \textit{Landmark Search with Image / Language}. For all of these tasks, the agent's final position is evaluated by measuring the Euclidean distance to the target location, and the prediction is considered correct if it falls within a threshold of $\varepsilon$. For the distance threshold $\varepsilon$, we set $\varepsilon = 10\,\text{m}$ for all tasks except \textit{Map Navigation}, where a larger threshold of $\varepsilon = 20\,\text{m}$ is used to account for noise caused by the marker's size on the map. A threshold-sensitivity check showed stable model rankings for $0.75\times$--$1.5\times$ thresholds (Kendall’s $\tau \geq 0.89$), a human–best LMM gap above 36 pp even at $\varepsilon=30$m, and Map Navigation changes within $\pm2$ pp for $\varepsilon$ = 15–25 m.

\noindent\textbf{4.~mean\_relative\_accuracy (MRA).}
MRA is a flexible evaluation metric applied to tasks that involve numerical estimation~\cite{yang25vsibench}. In our benchmark, this corresponds to the \textit{Object Count} task.
MRA averages the relative accuracy across a range of evaluation thresholds $\mathcal{C}=\{0.5, 0.55, \dots, 0.95\}$:

\begin{equation}
    \mathcal{MRA} = \frac{1}{10}\sum_{\theta \in \mathcal{C}}\mathbbm{1} \left( \frac{|\hat{y} - y|}{y} < 1-\theta \right),
\end{equation}
where $\hat{y}$ denotes the predicted value and $y$ the ground truth. 

MRA accounts for the magnitude of error, so it provides a more appropriate evaluation than a simple binary correct/incorrect classification.

\label{subsec:bench_evaluation}

\section{Navigation Agents}

\subsection{Problem Formulation}
The environment-agent interaction can be modeled as a partially observable sequential decision process: $\mathcal{E}=(S,A,\Omega,T)$, where $S$ represents the set of states, $A$ represents the set of actions, $\Omega$ represents the set of observations. The transition function is defined as $T:S\times A\rightarrow S$, with deterministic transitions between states conditioned on actions. At each time step $t$, the environment is in some state $s_t$ (\eg~a specific position and viewing direction). The agent receives a partial observation $o_t\in\Omega$, which consists of the visual input captured at time $t$ and, optionally, the current positional information shown as an image with a marker indicating the agent's location and orientation on the map. The agent also maintains a memory buffer $M_t \in \mathcal{M}$ that stores important information from previous steps up to $t-1$.
The agent then issues an action $a_t\in A$ conditioned on both $o_t$ and the stored memory $M_t$, which results in a new state $s_{t+1}\in S$ and a new observation $o_{t+1}\in\Omega$ from the updated viewpoint.
Simultaneously, relevant information from $o_t$ and thoughts is written to the memory, updating it to $M_{t+1}$.

\subsection{Baseline Agents}
For our experiments, we evaluate multiple LMMs as baseline agents: GPT-5~\cite{openai2025_gpt5_systemcard}, Claude Sonnet 4.5 (20250929)~\cite{anthropic2025_sonnet45_systemcard}, Gemini 2.5 Flash~\cite{comanici2025gemini25pushingfrontier}, Qwen2.5-VL-32B-Instruct~\cite{bai2025qwen25vltechnicalreport}, and InternVL3.5-8B and InternVL3.5-38B~\cite{wang2025internvl35advancingopensourcemultimodal}. Among these models, GPT-5, Claude Sonnet 4.5 and Gemini 2.5 Flash are closed-source, while Qwen2.5-VL-32B-Instruct, InternVL3.5-8B, and InternVL3.5-38B are open-source.
For inference with the open-source models, we used eight NVIDIA A100 80GB GPUs.
In addition, all tasks were executed in Unity 6000.0.30f1 (Unity 6 LTS) running on macOS.

In our setting, the action space $A$ consists of seven discrete actions: moving forward, tilting the viewpoint upward, tilting the viewpoint downward, rotating the viewpoint to the right, rotating the viewpoint to the left, resetting the viewpoint to align with the current heading direction, and outputting an answer. At branching points, the action space is augmented with movement actions corresponding to the available traversable directions, such as going forward, taking the right branch, or taking the left branch.

\label{subsec:baseline_agents}

\section{Experiment}
\subsection{Experimental Results}

\textbf{Human Performance.} 
We conducted human evaluations after obtaining approval from our institution's Institutional Review Board (IRB). We recruited five participants, including both undergraduate and graduate students. We selected participants who had been to Akihabara before or who visit the area regularly. Since real-world deployment in a specific urban district can benefit from local familiarity, we report this result as a local-expert human baseline, which serves as an in-domain upper-bound reference rather than a generic human baseline.

As shown in \Cref{table:overall_results}, human participants achieved higher accuracy than current LMMs. For tasks such as \textit{Map Navigation}, \textit{Landmark Search with Image}, \textit{Vision-Language Navigation}, and \textit{Relational Spatial Reasoning}, humans reached around 90\% accuracy. 
In contrast, performance on \textit{Localization}, \textit{Landmark Search with Language}, and \textit{Object Count} was more modest, at 68\%, 64\%, and 45\%, respectively.
In Landmark Search, the task configurations for the Language and Image variants are identical except for the input modality.
The fact that participants achieved much higher accuracy in the Image condition indicates that visual inputs contain substantially richer information than textual descriptions, including cues about appearance, location, and overall scene context.

\begin{table*}[!t]
  \centering
  \caption{\textbf{Overall Results (\%).} Comparison of model and human performance across seven spatial and reasoning tasks, grouped into three major categories. Gemini 2.5 Flash achieves the highest overall performance, while the strongest model varies across individual tasks. All LMMs still fall far short of human performance.}
  \resizebox{\textwidth}{!}{
  \begin{tabular}{lccccccc}
  \toprule
   & \multicolumn{3}{c}{\textbf{Environment Understanding}} 
   & \multicolumn{2}{c}{\textbf{Path Reasoning}} 
   & \multicolumn{2}{c}{\textbf{Spatial Reasoning}} \\
  \cmidrule(lr){2-4}
  \cmidrule(lr){5-6}
  \cmidrule(lr){7-8}
   & Loc & Landmark (Lang) & Landmark (Img) 
   & Map Nav & VLN 
   & Obj Count & Rel Reason \\
  \midrule
    GPT-5 & 8.0 & 16.0 & \textbf{48.0} & 0.0 & 8.0 & 2.4 & \textbf{32.0} \\
    Claude Sonnet 4.5 & 4.0 & 4.0 & 16.0 & \textbf{4.0} & 4.0 & 10.8 & 8.0 \\
    Gemini 2.5 Flash & \textbf{12.0} & \textbf{28.0} & 36.0 & 0.0 & 8.0 & \textbf{24.0} & 12.0 \\
    Qwen2.5-VL-32B-Instruct & 4.0 & 16.0 & 20.0 & 0.0 & 0.0 & 18.8 & 4.0 \\
    InternVL3.5-8B & 4.0 & 20.0 & 20.0 & 0.0 & \textbf{12.0} & 2.8 & 0.0 \\
    InternVL3.5-38B & 0.0 & 16.0 & 0.0 & 0.0 & \textbf{12.0} & 7.2 & 4.0 \\
  \midrule
  Human & 68.0 & 64.0 & 92.0 & 92.0 & 88.0 & 45.2 & 92.0 \\
  \bottomrule
  \end{tabular}
  }
  \label{table:overall_results}
\end{table*}
  
\begin{figure*}[!t]
  \centering
      \includegraphics[width=0.95\linewidth]{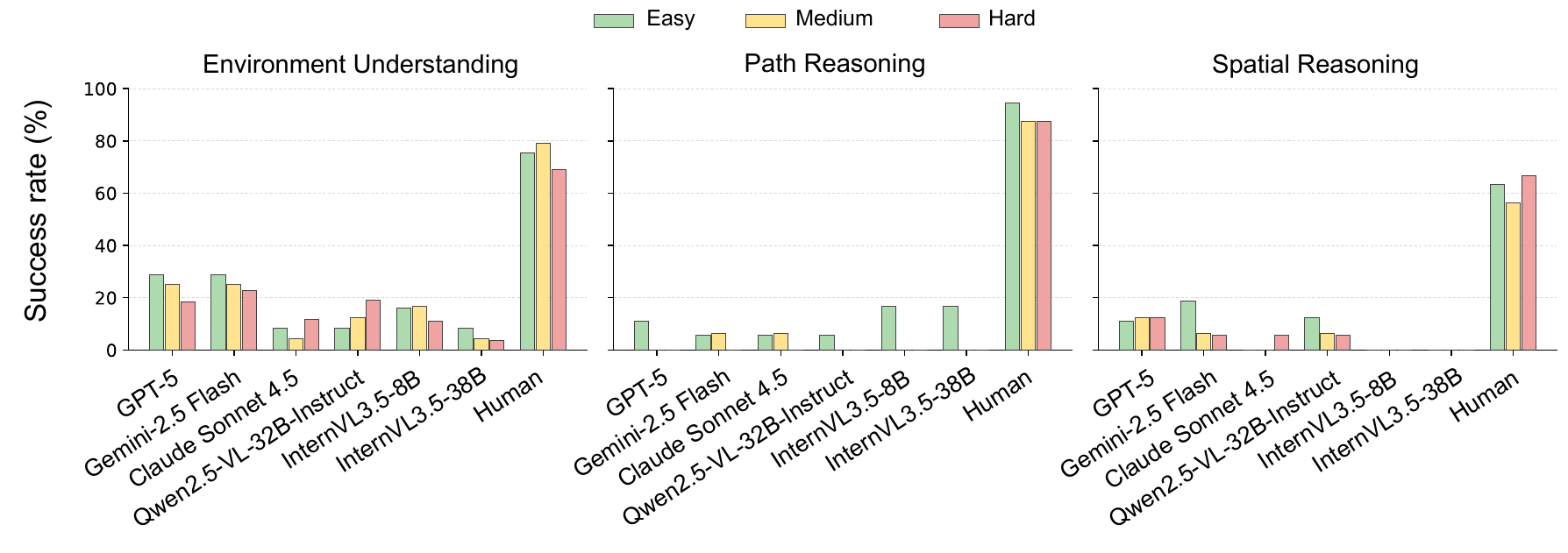}\\
     \caption{\textbf{Success rate by difficulty level across tasks and models (\%).} This graph shows model and human performance for each task, divided into three difficulty levels: Easy (E), Medium (M), and Hard (H).}
      \label{fig:result_by_difficulty}
\end{figure*}

Next, our main findings are as follows:

\noindent\textbf{F1: LMMs perform far below human level.}
As shown in \Cref{table:overall_results}, all LMMs fall far short of human performance across the benchmark. Gemini 2.5 Flash achieves the highest overall performance among the evaluated models, while the strongest model varies across individual tasks (e.g., GPT-5 on \textit{Landmark (Img)} and \textit{Rel Reason}). These results indicate that, despite incremental progress, substantial gaps to human-level environment understanding and spatial reasoning remain.

\vspace{2mm}
\noindent\textbf{F2: Image-based landmark search can be easier than language-based search.}
Across models, \textit{Landmark Search with Image} generally outperforms \textit{Landmark Search with Language} (e.g., GPT-5: 48.0 vs.\ 16.0; Claude: 16.0 vs.\ 4.0; Qwen: 20.0 vs.\ 16.0), suggesting that visual inputs provide concrete cues (appearance, textures, and surrounding context) that more directly support navigation than abstract textual descriptions. However, this improvement is not universal. InternVL does not show a clear gain from image inputs (8B: 20.0 vs.\ 20.0; 38B: 0.0 vs.\ 16.0), indicating limitations in effectively leveraging landmark images. By including both image- and language-based variants, our benchmark enables analysis of how models leverage different input modalities.

\vspace{2mm}
\noindent\textbf{F3: Performance decreases as task difficulty increases.}
As shown in \Cref{fig:result_by_difficulty}, model performance generally declines with higher task difficulty. For example, GPT-5's accuracy in Environment Understanding drops from 28.0 → 25.0 → 18.5, and in Path Reasoning from 11.1 → 0.0 → 0.0 across the Easy, Medium, and Hard settings.
This demonstrates that the benchmark allows meaningful comparison of model performance across different difficulty levels.

\begin{table*}[!t]
  \centering
  \caption{\textbf{Comparison with and without location information.} 
  We observe that the performance did not improve consistently across tasks; in some cases, accuracy decreased instead.
  } 
  \resizebox{\textwidth}{!}{
  \begin{tabular}{lccccccc}
  \toprule
   & \multicolumn{3}{c}{\textbf{Environment Understanding}} 
   & \multicolumn{2}{c}{\textbf{Path Reasoning}} 
   & \multicolumn{2}{c}{\textbf{Spatial Reasoning}} \\
  \cmidrule(lr){2-4}
  \cmidrule(lr){5-6}
  \cmidrule(lr){7-8}
   & Loc & Landmark (Lang) & Landmark (Img) 
   & Map Nav & VLN 
   & Obj Count & Rel Reason \\
  \midrule
  GPT-5 & 8.0 & 16.0 & 48.0 & 0.0 & 8.0 & 2.4 & 32.0 \\
  GPT-5 (w/o location) & - & 24.0 & 44.0 & 4.0 & 8.0 & 0.0 & 48.0 \\
  \bottomrule
  \end{tabular}
  }
  \label{table:comparison_location}
  \end{table*}
  
  \begin{figure*}[!t]
  \centering
      \includegraphics[width=0.99\linewidth]{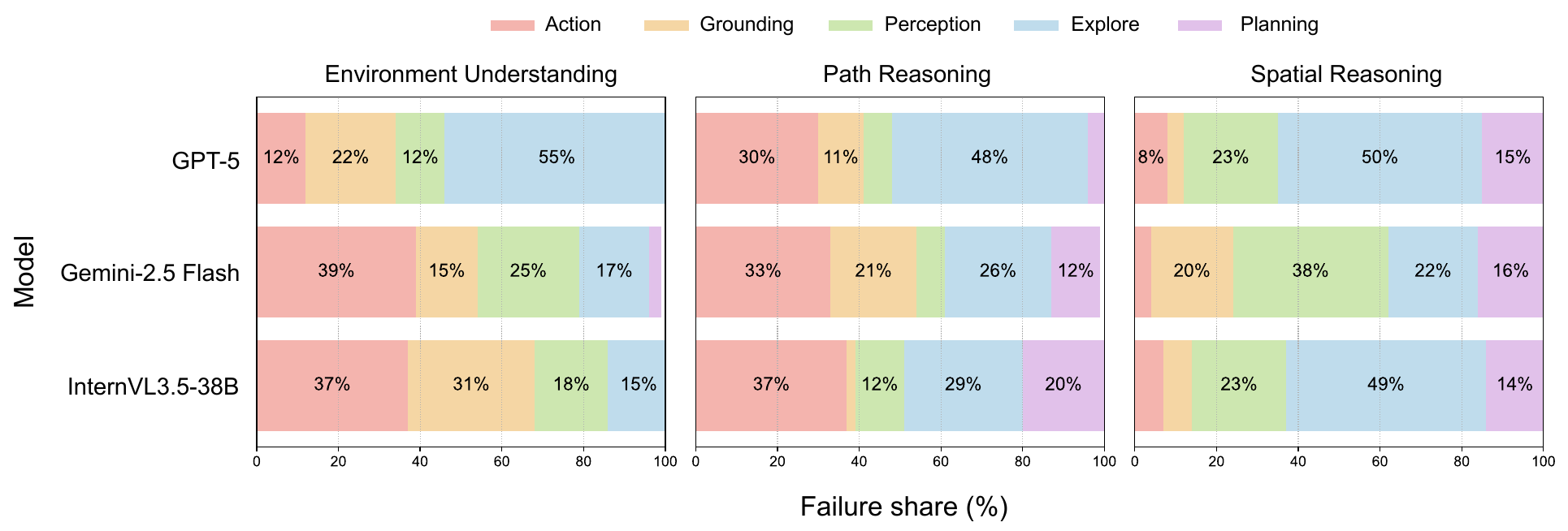}\\
     \caption{\textbf{Failure cause breakdown by task category across models (\%).} This figure shows the distribution of failure causes for each model across three task categories. Failures are categorized into five types: Action, Grounding, Perception, Explore, and Planning, and each stacked bar reports the percentage breakdown within the corresponding (model, task category) setting.}
      \label{fig:failure_stackedbar}
  \end{figure*}

\subsection{Analysis}

\vspace{2mm}
\noindent\textbf{Effect of Location Information.}
As shown in \Cref{table:comparison_location}, we examined how explicitly providing the agent with its current location affects its spatial understanding and reasoning abilities, and found that adding a location prior yields inconsistent performance improvements across tasks. In Environment Understanding, Landmark (Language) accuracy dropped from 24.0\% to 16.0\%, while Landmark (Image) showed no meaningful change (within the margin of error). Path Reasoning tasks (Map Navigation and VLN) exhibited nearly no difference, suggesting static location information fails to aid path planning or instruction interpretation. In Spatial Reasoning, Object Count improved slightly, but Relational Reasoning fell from 48.0\% to 32.0\%. These performance degradations suggest agents struggle to align map-based location and relational information with real-world cues derived from visual inputs, rather than location information being inherently unhelpful. While prior work addresses map-based reasoning \cite{xing2025map, paz-argaman19run, schumann-riezler-2021-generating}, the impact of the map-to-visual alignment process remains under-examined. Future research should independently evaluate this alignment capability and, where necessary, provide mechanisms for learning it.

\vspace{2mm}
\noindent\textbf{Distinct Failure Patterns across Models.}
To better understand failure factors for three major models (GPT-5, Gemini 2.5 Flash, and InternVL3.5-38B), we used Gemini 3 Flash to analyze failures from viewpoint images, maps, and thought histories, assigning each case to one of five categories—Action, Grounding, Perception, Explore, and Planning—and then had humans verify the label and select the single dominant cause. Action denotes low-level action errors (\eg, overshoot, wrong action, premature answer); Grounding, self-localization/orientation inconsistencies; Perception, object-recognition failures; Explore, stagnation or loops; and Planning, missed instructions or requirements. The resulting breakdown by model and task category is shown in \Cref{fig:failure_stackedbar}. These results should be interpreted as diagnosing integrated embodied urban performance, rather than isolating pure spatial reasoning: Action and Explore failures indicate that current LMM agents also suffer from controller and prompting brittleness when converting perception and reasoning into navigation decisions.

The results reveal distinct failure patterns reflecting model capabilities. First, GPT-5 contrasts sharply with other models: it shows minimal Action failures (12\% in Environment Understanding) but dominant Explore failures (55\%), indicating effective basic execution but poor exploration strategy. Conversely, Gemini 2.5 Flash and InternVL3.5-38B struggle primarily with Action failures (around 40\%), pointing to deficits in low-level control. Second, Perception failures surge in Spatial Reasoning across all models (reaching 38\% for Gemini), confirming that overlooking minute visual details is fatal for spatial tasks. Furthermore, Gemini displays consistent Grounding errors (15–21\%) across categories, indicating persistent difficulty in aligning map data with the first-person perspective.

\vspace{2mm}
\noindent\textbf{Case Study of Successes and Failures.}
\begin{figure}[t]
  \centering
  \begin{subfigure}[t]{0.45\columnwidth}
    \centering
    \includegraphics[width=\linewidth]{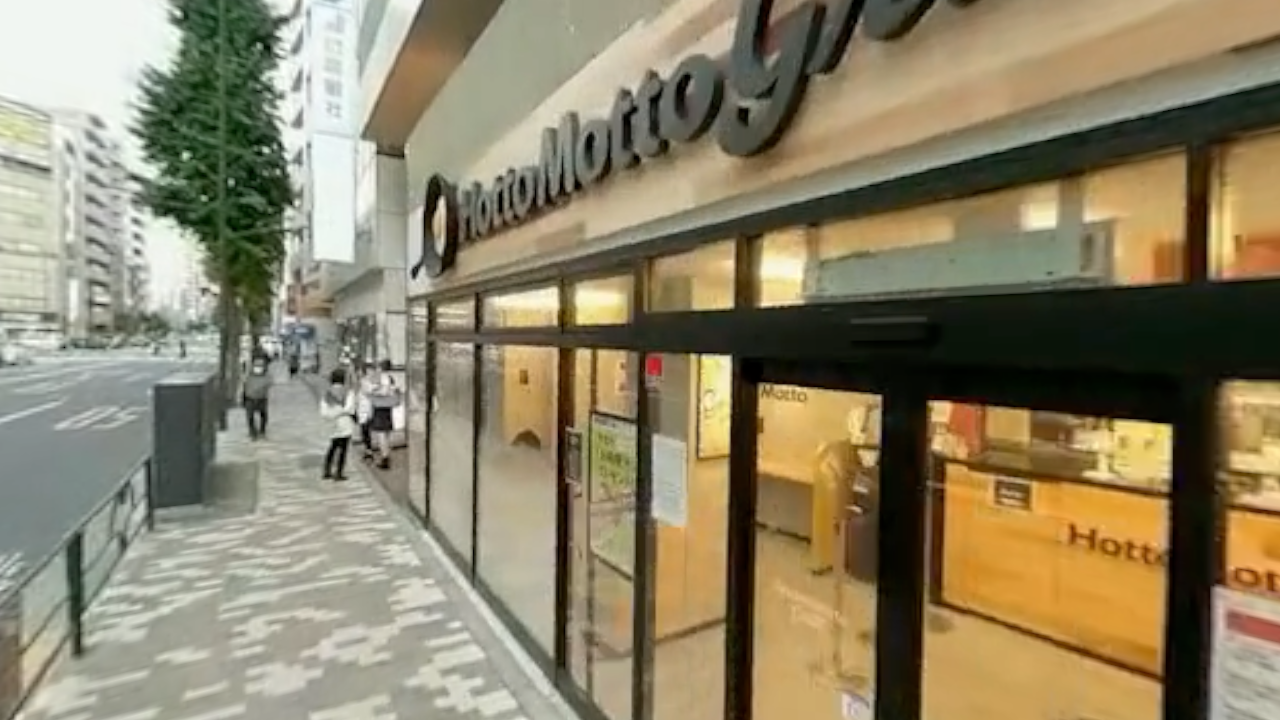}
    \caption*{t = 7. HottoMotto is on the right.}
  \end{subfigure}\hfill
  \begin{subfigure}[t]{0.45\columnwidth}
    \centering
    \includegraphics[width=\linewidth]{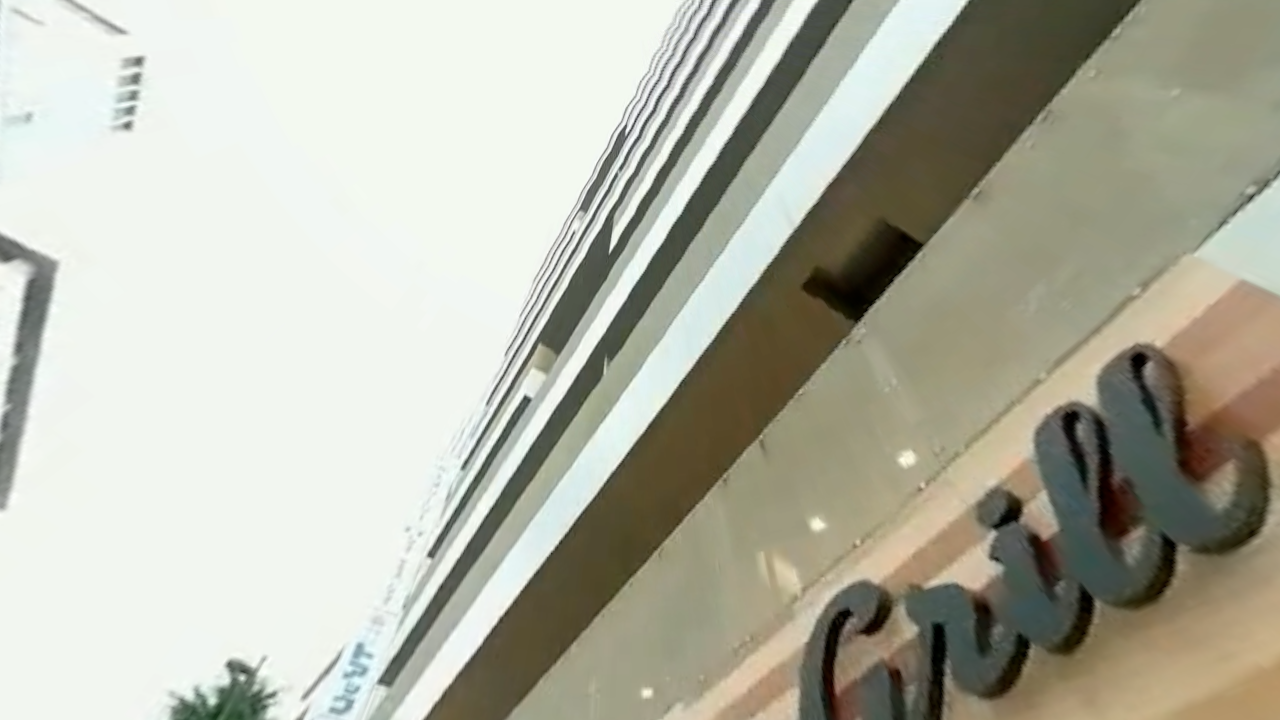}
    \caption*{t = 8. Look above.}
  \end{subfigure}

  \vspace{0.6em}

  \begin{subfigure}[t]{0.45\columnwidth}
    \centering
    \includegraphics[width=\linewidth]{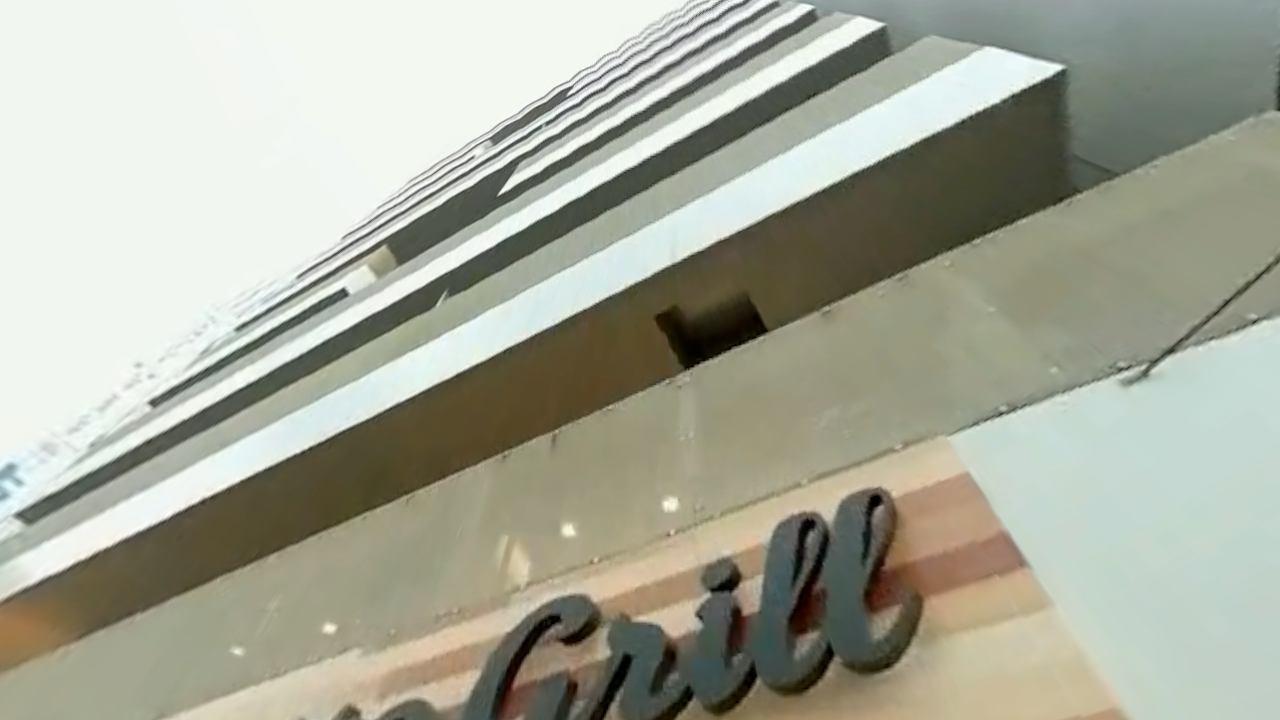}
    \caption*{t = 40. Look to the right.}
  \end{subfigure}\hfill
  \begin{subfigure}[t]{0.45\columnwidth}
    \centering
    \includegraphics[width=\linewidth]{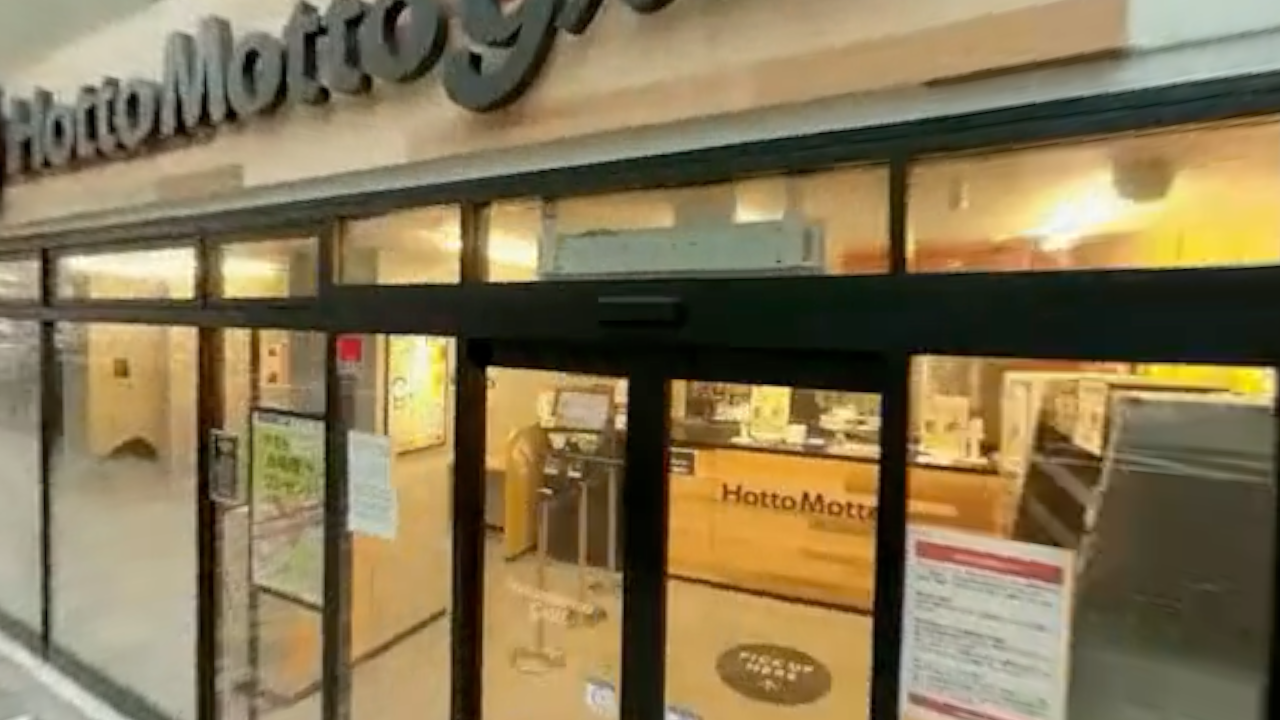}
    \caption*{t = 41. Look below.}
  \end{subfigure}

  \caption{\textbf{Example of the agent's views for \textit{Landmark Search with Language}, resulting in failure.} The agent is instructed to search for ``Jonathan''. From t = 8 to t = 38, the gaze repeatedly moved up and down. At t = 39, the gaze briefly shifted to the right, but from t = 41 onward, 
  it returned to the same up–down movement.}
  \label{fig:indiv_failure}
\end{figure}
For Landmark Search tasks in which the Language variant fails but the Image variant succeeds, we analyze individual cases using GPT-5, examining the agent's visual observations. 
The example shown here is a task in which the agent must locate Jonathan's restaurant.

In the \textit{Landmark Search with Language} failure case in Figure~\ref{fig:indiv_failure}, the agent passes by Hotto Motto and hypothesizes that Jonathan is located above it, prompting it to look upward. After failing to find the target, it shifts its gaze right and continues scanning. As a result, the agent stays in the same location, repeatedly changing viewpoints until the episode ends due to the step limit.

\begin{figure}[t]
  \centering
  \begin{subfigure}[t]{0.45\columnwidth}
    \centering
    \includegraphics[width=\linewidth]{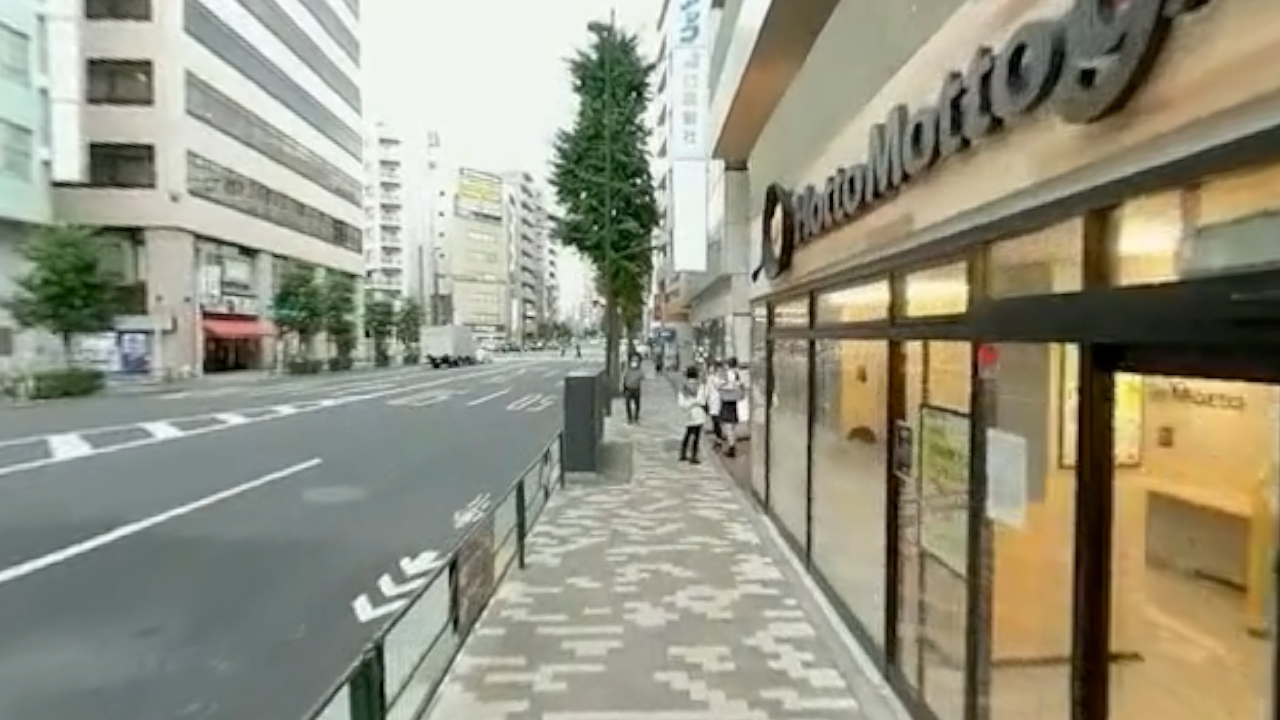}
    \caption*{t = 4. HottoMotto is on the right.}
  \end{subfigure}\hfill
  \begin{subfigure}[t]{0.45\columnwidth}
    \centering
    \includegraphics[width=\linewidth]{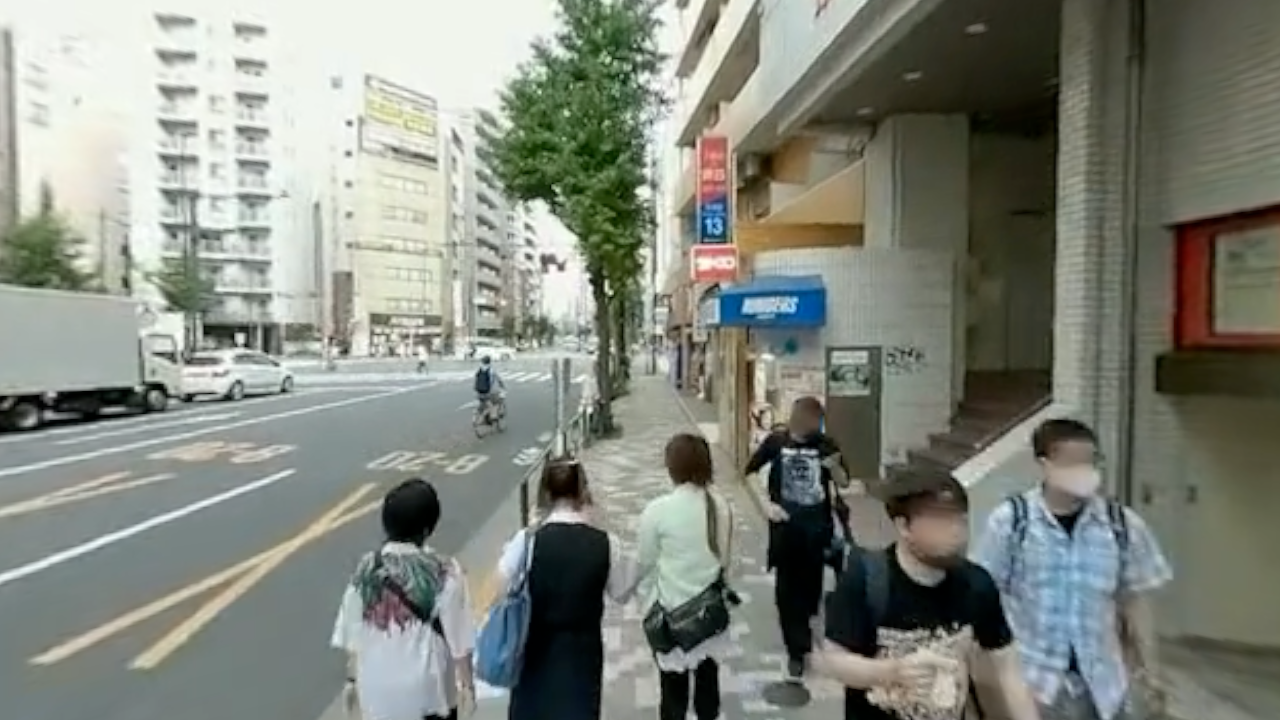}
    \caption*{t = 5. Go past.}
  \end{subfigure}

  \vspace{0.6em}

  \begin{subfigure}[t]{0.45\columnwidth}
    \centering
    \includegraphics[width=\linewidth]{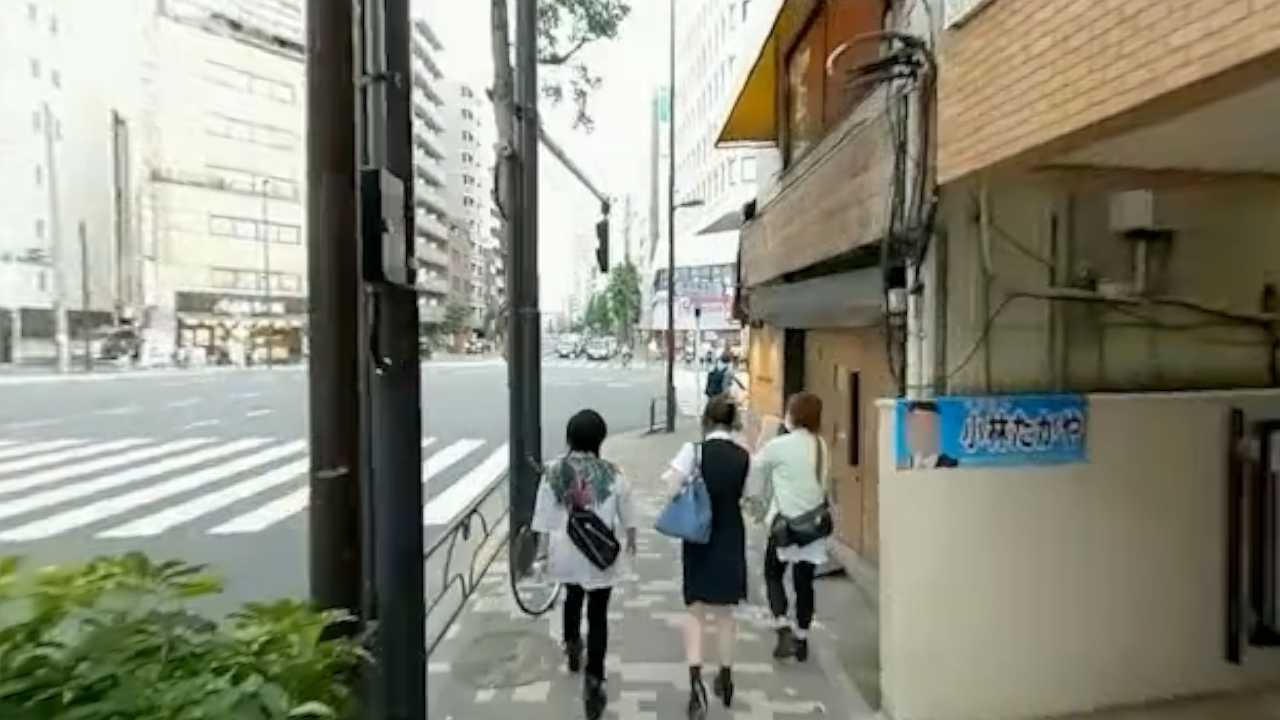}
    \caption*{t = 16. Spot blue and white vertical stripes.}
  \end{subfigure}\hfill
  \begin{subfigure}[t]{0.45\columnwidth}
    \centering
    \includegraphics[width=\linewidth]{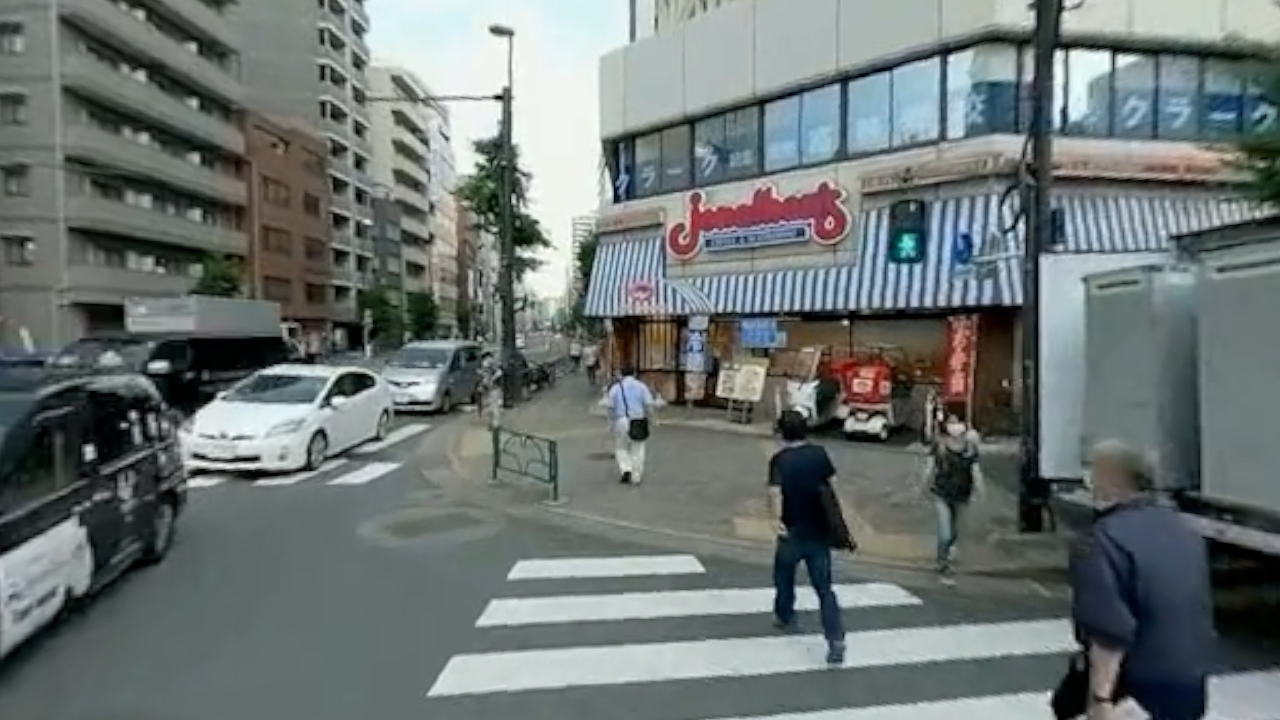}
    \caption*{t = 29. Arrive at the destination.}
  \end{subfigure}

  \caption{\textbf{Example of the agent's views for \textit{Landmark Search with Image}, resulting in success.} The agent is instructed to search for ``Jonathan'' given in image form.}
  \label{fig:indiv_success}
\end{figure}

In contrast, as shown in Figure~\ref{fig:indiv_success}, in \textit{Landmark Search with Image}, the target image provides cues such as its street-corner location and distinctive blue-and-white vertical facade with a red sign. The agent therefore disregards Hotto Motto, moves toward the major intersection, and successfully finds the target building.

\section{Limitations and Future Work}

\vspace{2mm}
\noindent\textbf{Restricted Exploration and Limited Interaction.}
Because 360CityArena is constructed from pre-recorded 360\textdegree{}~videos, agents can explore only along captured trajectories rather than move freely to arbitrary locations. As a result, transitions across trajectory boundaries may introduce discontinuities that would not arise in fully interactive 3D simulators or in the real world. In our logs, boundary-transition actions accounted for 11.3\% of all actions and were not overrepresented among Action or Explore failures; manual inspection also did not reveal clear discontinuity-induced failures. Nonetheless, we argue that situated visual navigation and spatial reasoning in a highly photorealistic urban setting remain challenging and important for current LMM-based agents. Extending the environment to support freer exploration and richer interaction is an important direction for future work.

\vspace{2mm}
\noindent\textbf{Limited Urban Diversity.}
A limitation of the current benchmark is its restriction to a single urban district, which may limit urban diversity and conclusions regarding cross-regional generalization. However, we consider that 360CityArena provides a uniquely photorealistic, large-scale setting that facilitates more realistic urban evaluations than existing benchmarks. We leave cross-regional generalization as an important direction for future work.

\section{Conclusions}
We present 360CityArena, a benchmark designed to evaluate the urban exploration capabilities of embodied agents within a photorealistic environment constructed from 360-degree videos. Our experimental results reveal that state-of-the-art LMM-based agents still face significant challenges in understanding and reasoning within city environments. We believe that providing a benchmark closer to real-world conditions will further advance the development of embodied agents that are applicable to real urban settings.

\section*{Acknowledgements}
This work was supported in part by JSPS KAKENHI Grant Number 25H01164, SIP Smart Disaster Prevention, JST BOOST Grant Number JPMJBS2418, and JST ASPIRE Program Grant No.~JPMJAP2303. This study was conducted with approval from the relevant Ethics Committee (Approval No. UT-IST-RE-250508\_6).

\newpage
%
%
\bibliographystyle{splncs04}
\bibliography{arxiv}

@String(IJCV  = {Int. J. Comput. Vis.})

@String(CVPR  = {IEEE Conf. Comput. Vis. Pattern Recog.})

@String(ICCV  = {Int. Conf. Comput. Vis.})

@String(ECCV  = {Eur. Conf. Comput. Vis.})

@String(NeurIPS = {Adv. Neural Inform. Process. Syst.})

@String(ICML  = {Int. Conf. Mach. Learn.})

@String(ICLR  = {Int. Conf. Learn. Represent.})

@String(AAAI  = {AAAI})

@String(TOG   = {ACM Trans. Graph.})

@String(IJCV  = {IJCV})

@String(CVPR  = {CVPR})

@String(ICCV  = {ICCV})

@String(ECCV  = {ECCV})

@String(NeurIPS = {NeurIPS})

@String(ICML  = {ICML})

@String(ICLR  = {ICLR})

@String(TOG   = {ACM TOG})

@inproceedings{wu25metaurban,
  title     = {{MetaUrban}: An Embodied {AI} Simulation Platform for Urban Micromobility},
  author    = {Wayne Wu and Honglin He and Jack He and Yiran Wang and Chenda Duan and Zhizheng Liu and Quanyi Li and Bolei Zhou},
  booktitle = {ICLR},
  year      = {2025}
}

@inproceedings{yang25virl,
  author    = {Yang, Jihan
               and Ding, Runyu
               and Brown, Ellis
               and Qi, Xiaojuan
               and Xie, Saining},
  title     = {{V-IRL}: Grounding Virtual Intelligence in Real Life},
  booktitle = {ECCV},
  year      = {2024},
  volume = {15103},
  pages = {36--55}
}

@inproceedings{feng25citybench,
  title     = {{CityBench}: Evaluating the Capabilities of Large Language Models for Urban Tasks},
  author    = {Feng, Jie and Zhang, Jun and Liu, Tianhui and Zhang, Xin and Ouyang, Tianjian and Yan, Junbo and Du, Yuwei and Guo, Siqi and Li, Yong},
  booktitle = {KDD},
  year      = {2025},
  pages = {5413--5424}
}

@inproceedings{paz-argaman19run,
  title     = {{RUN} through the Streets: A New Dataset and Baseline Models for Realistic Urban Navigation},
  author    = {Paz-Argaman, Tzuf  and
               Tsarfaty, Reut},
  booktitle = {EMNLP-IJCNLP},
  year      = {2019},
  pages = {6449--6455}
}

@inproceedings{sugimoto2020moviemap,
  title     = {Building {Movie Map} -- A Tool for Exploring Areas in a City -- and its Evaluations},
  author    = {Sugimoto, Naoki and Ebine, Yoshihito and Aizawa, Kiyoharu},
  booktitle = {ACM MM},
  pages     = {3330--3338},
  year      = {2020}
}

@inproceedings{lee2025citynav,
  author    = {Lee, Jungdae and Miyanishi, Taiki and Kurita, Shuhei and Sakamoto, Koya and Azuma, Daichi and Matsuo, Yutaka and Inoue, Nakamasa},
  title     = {{CityNav}: A Large-Scale Dataset for Real-World Aerial Navigation},
  booktitle = {ICCV},
  year      = {2025},
  pages     = {5912--5922}
}

@inproceedings{zhang2024tagmap,
  title     = {{Tag Map}: A Text-Based Map for Spatial Reasoning and Navigation with Large Language Models},
  author    = {Mike Zhang and Kaixian Qu and Vaishakh Patil and Cesar Cadena and Marco Hutter},
  booktitle = {CoRL},
  year      = {2024},
  volume = {270},
  pages = {2120--2146}
}

@inproceedings{mukhopadhyay2025mapwise,
  title     = {{MAPW}ise: Evaluating Vision-Language Models for Advanced Map Queries},
  author    = {Mukhopadhyay, Srija  and
               Rajgaria, Abhishek  and
               Khatiwada, Prerana  and
               Shrivastava, Manish  and
               Roth, Dan  and
               Gupta, Vivek},
  booktitle = {NAACL},
  year      = {2025},
  pages     = {9348--9378}
}

@article{takenawa2025building,
  title   = {Building and Evaluating a Realistic Virtual World for Large Scale Urban Exploration from 360\textdegree{} Videos},
  author  = {Takenawa, Mizuki and Sugimoto, Naoki and W{\"o}hler, Leslie and Ikehata, Satoshi and Aizawa, Kiyoharu},
  journal = {MTAP},
  volume = {85},
  eid = {149},
  year = {2026}
}

@inproceedings{yang25vsibench,
  author    = {Yang, Jihan and Yang, Shusheng and Gupta, Anjali W. and Han, Rilyn and Fei-Fei, Li and Xie, Saining},
  title     = {Thinking in Space: How Multimodal Large Language Models See, Remember, and Recall Spaces},
  booktitle = {CVPR},
  year      = {2025},
  pages     = {10632--10643}
}

@inproceedings{zhou2023webarena,
  title     = {{WebArena}: A Realistic Web Environment for Building Autonomous Agents},
  author    = {Zhou, Shuyan and Xu, Frank F and Zhu, Hao and Zhou, Xuhui and Lo, Robert and Sridhar, Abishek and Cheng, Xianyi and Ou, Tianyue and Bisk, Yonatan and Fried, Daniel and others},
  booktitle = {ICLR},
  year      = {2024}
}

@article{miyai2025webchorearena,
  title   = {{WebChoreArena}: Evaluating Web Browsing Agents on Realistic Tedious Web Tasks},
  author  = {Miyai, Atsuyuki and Zhao, Zaiying and Egashira, Kazuki and Sato, Atsuki and Sunada, Tatsumi and Onohara, Shota and Yamanishi, Hiromasa and Toyooka, Mashiro and Nishina, Kunato and Maeda, Ryoma and others},
  journal = {arXiv preprint arXiv:2506.01952},
  year    = {2025}
}

@inproceedings{mirowski2018learning,
  title   = {Learning to Navigate in Cities Without a Map},
  author  = {Mirowski, Piotr and Grimes, Matt and Malinowski, Mateusz and Hermann, Karl Moritz and Anderson, Keith and Teplyashin, Denis and Simonyan, Karen and Kavukcuoglu, Koray and Zisserman, Andrew and Hadsell, Raia},
  booktitle = {NeurIPS},
  volume  = {31},
  year    = {2018}
}

@article{comanici2025gemini25pushingfrontier,
  title   = {{Gemini 2.5}: Pushing the Frontier with Advanced Reasoning, Multimodality, Long Context, and Next Generation Agentic Capabilities},
  author  = {Gheorghe Comanici and others},
  year    = {2025},
  journal = {arXiv preprint arXiv:2507.06261}
}

@article{bai2025qwen25vltechnicalreport,
  title   = {{Qwen2.5-VL} Technical Report},
  author  = {Shuai Bai and Keqin Chen and Xuejing Liu and Jialin Wang and Wenbin Ge and Sibo Song and Kai Dang and Peng Wang and Shijie Wang and Jun Tang and Humen Zhong and Yuanzhi Zhu and Mingkun Yang and Zhaohai Li and Jianqiang Wan and Pengfei Wang and Wei Ding and Zheren Fu and Yiheng Xu and Jiabo Ye and Xi Zhang and Tianbao Xie and Zesen Cheng and Hang Zhang and Zhibo Yang and Haiyang Xu and Junyang Lin},
  year    = {2025},
  journal = {arXiv preprint arXiv:2502.13923}
}

@article{wang2025internvl35advancingopensourcemultimodal,
  title   = {{InternVL3.5}: Advancing Open-Source Multimodal Models in Versatility, Reasoning, and Efficiency},
  author  = {Weiyun Wang and Zhangwei Gao and Lixin Gu and Hengjun Pu and Long Cui and Xingguang Wei and Zhaoyang Liu and Linglin Jing and Shenglong Ye and Jie Shao and Zhaokai Wang and Zhe Chen and Hongjie Zhang and Ganlin Yang and Haomin Wang and Qi Wei and Jinhui Yin and Wenhao Li and Erfei Cui and Guanzhou Chen and Zichen Ding and Changyao Tian and Zhenyu Wu and Jingjing Xie and Zehao Li and Bowen Yang and Yuchen Duan and Xuehui Wang and Zhi Hou and Haoran Hao and Tianyi Zhang and Songze Li and Xiangyu Zhao and Haodong Duan and Nianchen Deng and Bin Fu and Yinan He and Yi Wang and Conghui He and Botian Shi and Junjun He and Yingtong Xiong and Han Lv and Lijun Wu and Wenqi Shao and Kaipeng Zhang and Huipeng Deng and Biqing Qi and Jiaye Ge and Qipeng Guo and Wenwei Zhang and Songyang Zhang and Maosong Cao and Junyao Lin and Kexian Tang and Jianfei Gao and Haian Huang and Yuzhe Gu and Chengqi Lyu and Huanze Tang and Rui Wang and Haijun Lv and Wanli Ouyang and Limin Wang and Min Dou and Xizhou Zhu and Tong Lu and Dahua Lin and Jifeng Dai and Weijie Su and Bowen Zhou and Kai Chen and Yu Qiao and Wenhai Wang and Gen Luo},
  year    = {2025},
  journal = {arXiv preprint arXiv:2508.18265}
}

@techreport{anthropic2025_sonnet45_systemcard,
  author       = {{Anthropic}},
  title        = {{Claude Sonnet 4.5} System Card},
  institution  = {Anthropic, PBC},
  year         = {2025},
  howpublished = {\url{https://www.anthropic.com/claude-sonnet-4-5-system-card}}
}

@techreport{openai2025_gpt5_systemcard,
  author       = {{OpenAI}},
  title        = {{GPT-5} System Card},
  institution  = {OpenAI},
  year         = {2025},
  howpublished = {\url{https://cdn.openai.com/gpt-5-system-card.pdf}}
}

@article{ji2025towards,
  title   = {Towards Autonomous {UAV} Visual Object Search in City Space: Benchmark and Agentic Methodology},
  author  = {Ji, Yatai and Zhu, Zhengqiu and Zhao, Yong and Liu, Beidan and Gao, Chen and Zhao, Yihao and Qiu, Sihang and Hu, Yue and Yin, Quanjun},
  journal = {AAAI},
  year    = {2026},
  volume = {40},
  number = {22},
  pages = {18342--18350}
}

@article{xing2025map,
  title   = {Can Large Vision Language Models Read Maps Like a Human?},
  author  = {Shuo Xing and Zezhou Sun and Shuangyu Xie and Kaiyuan Chen and Yanjia Huang and Yuping Wang and Jiachen Li and Dezhen Song and Zhengzhong Tu},
  journal = {arXiv preprint arXiv:2503.14607},
  year    = {2025}
}

@article{anderson2018evaluation,
  title   = {On Evaluation of Embodied Navigation Agents},
  author  = {Anderson, Peter and Chang, Angel and Chaplot, Devendra Singh and Dosovitskiy, Alexey and Gupta, Saurabh and Koltun, Vladlen and Kosecka, Jana and Malik, Jitendra and Mottaghi, Roozbeh and Savva, Manolis and others},
  journal = {arXiv preprint arXiv:1807.06757},
  year    = {2018}
}

@article{duan2022survey,
  author  = {Duan, Jiafei and Yu, Samson and Tan, Hui Li and Zhu, Hongyuan and Tan, Cheston},
  journal = {TETCI},
  title   = {A Survey of Embodied {AI}: From Simulators to Research Tasks},
  year    = {2022},
  volume  = {6},
  number = {2},
  pages   = {230--244}
}

@article{yang2025survey,
  author  = {Liu, Yang and Chen, Weixing and Bai, Yongjie and Liang, Xiaodan and Li, Guanbin and Gao, Wen and Lin, Liang},
  journal = {IEEE/ASME TMECH},
  title   = {Aligning Cyber Space With Physical World: A Comprehensive Survey on Embodied {AI}},
  year    = {2025},
  volume = {30},
  number = {6},
  pages = {7253--7274}
}

@inproceedings{das2018eqa,
  author    = {Das, Abhishek and Datta, Samyak and Gkioxari, Georgia and Lee, Stefan and Parikh, Devi and Batra, Dhruv},
  title     = {Embodied Question Answering},
  booktitle = {CVPR},
  year      = {2018},
  pages = {1--10}
}

@inproceedings{liu2025citywalker,
  title     = {{CityWalker}: Learning Embodied Urban Navigation from Web-Scale Videos},
  author    = {Liu, Xinhao and Li, Jintong and Jiang, Yicheng and Sujay, Niranjan and Yang, Zhicheng and Zhang, Juexiao and Abanes, John and Zhang, Jing and Feng, Chen},
  booktitle = {CVPR},
  year      = {2025},
  pages = {6875--6885}
}

@inproceedings{Chen_2019_CVPR,
  author    = {Chen, Howard and Suhr, Alane and Misra, Dipendra and Snavely, Noah and Artzi, Yoav},
  title     = {{TOUCHDOWN}: Natural Language Navigation and Spatial Reasoning in Visual Street Environments},
  booktitle = {CVPR},
  year      = {2019},
  pages = {12538--12547}
}

@article{li2024vln,
  title     = {{VLN-Video}: Utilizing Driving Videos for Outdoor Vision-and-Language Navigation},
  author    = {Li, Jialu and Padmakumar, Aishwarya and Sukhatme, Gaurav and Bansal, Mohit},
  journal = {AAAI},
  volume    = {38},
  number = {17},
  year      = {2024},
  pages = {18517--18526}
}

@inproceedings{jiao2025litevloc,
  title     = {{LiteVLoc}: Map-Lite Visual Localization for Image Goal Navigation},
  author    = {Jiao, Jianhao and He, Jinhao and Liu, Changkun and Aegidius, Sebastian and Hu, Xiangcheng and Braud, Tristan and Kanoulas, Dimitrios},
  booktitle = {ICRA},
  year      = {2025},
  pages = {5244--5251}
}

@inproceedings{hong2025embodied,
  title     = {Embodied Web Agents: Bridging Physical-Digital Realms for Integrated Agent Intelligence},
  author    = {Yining Hong and Rui Sun and Bingxuan Li and Xingcheng Yao and Maxine Wu and Alexander Chien and Da Yin and Ying Nian Wu and Zhecan Wang and Kai-Wei Chang},
  booktitle = {NeurIPS},
  year      = {2025},
  volume = {38},
}

@article{gao2024embodiedcity,
  title   = {{EmbodiedCity}: A Benchmark Platform for Embodied Agent in Real-World City Environment},
  author  = {Gao, Chen and Zhao, Baining and Zhang, Weichen and Mao, Jinzhu and Zhang, Jun and Zheng, Zhiheng and Man, Fanhang and Fang, Jianjie and Zhou, Zile and Cui, Jinqiang and others},
  journal = {arXiv preprint arXiv:2410.09604},
  year    = {2024}
}

@inproceedings{xie2025vid2sim,
  title     = {{Vid2Sim}: Realistic and Interactive Simulation from Video for Urban Navigation},
  author    = {Xie, Ziyang and Liu, Zhizheng and Peng, Zhenghao and Wu, Wayne and Zhou, Bolei},
  booktitle = {CVPR},
  year      = {2025},
  pages     = {1581--1591}
}

@inproceedings{Yang2023FreeNeRF,
  author    = {Jiawei Yang and Marco Pavone and Yue Wang},
  title     = {{FreeNeRF}: Improving Few-shot Neural Rendering with Free Frequency Regularization},
  booktitle = {CVPR},
  year      = {2023},
  pages     = {8254--8263}
}

@inproceedings{xu2024fewshotnerfadaptiverendering,
  title     = {Few-Shot {NeRF} by Adaptive Rendering Loss Regularization},
  author    = {Qingshan Xu and Xuanyu Yi and Jianyao Xu and Wenbing Tao and Yew-Soon Ong and Hanwang Zhang},
  booktitle = {ECCV},
  year      = {2024},
  pages = {125--142},
  volume = {15124}
}

@inproceedings{zhang2024gaussian,
  title     = {{Gaussian} in the Wild: {3D} {Gaussian} Splatting for Unconstrained Image Collections},
  author    = {Zhang, Dongbin and Wang, Chuming and Wang, Weitao and Li, Peihao and Qin, Minghan and Wang, Haoqian},
  booktitle = {ECCV},
  year      = {2024},
  volume = {15134},
  pages = {341--359}
}

@inproceedings{barron2022mipnerf360,
  title     = {{Mip-NeRF 360}: Unbounded Anti-Aliased Neural Radiance Fields},
  author    = {Jonathan T. Barron and Ben Mildenhall and 
               Dor Verbin and Pratul P. Srinivasan and Peter Hedman},
  booktitle = {CVPR},
  year      = {2022},
  pages     = {5470--5479}
}

@inproceedings{mildenhall2020nerf,
  title     = {{NeRF}: Representing Scenes as Neural Radiance Fields for View Synthesis},
  author    = {Ben Mildenhall and Pratul P. Srinivasan and Matthew Tancik and Jonathan T. Barron and Ravi Ramamoorthi and Ren Ng},
  year      = {2020},
  booktitle = {ECCV},
  volume = {12346},
  pages = {405--421}
}

@article{kerbl3Dgaussians,
  author  = {Kerbl, Bernhard and Kopanas, Georgios and Leimk{\"u}hler, Thomas and Drettakis, George},
  title   = {{3D} Gaussian Splatting for Real-Time Radiance Field Rendering},
  journal = {ACM TOG},
  year    = {2023},
  volume = {42},
  number = {4},
  eid = {139},
  pages = {1--14}
}

@inproceedings{brahmbhatt2017deepnav,
  author    = {Brahmbhatt, Samarth and Hays, James},
  title     = {{DeepNav}: Learning to Navigate Large Cities},
  booktitle = {CVPR},
  year      = {2017},
  pages = {5193--5202}
}

@article{savva2017minos,
  author  = {Manolis Savva and Angel X. Chang and Alexey Dosovitskiy and Thomas Funkhouser and Vladlen Koltun},
  title   = {{MINOS}: Multimodal Indoor Simulator for Navigation in Complex Environments},
  journal = {arXiv preprint  arXiv:1712.03931},
  year    = {2017}
}

@inproceedings{savva2019habitat,
  author    = {Savva, Manolis and Kadian, Abhishek and Maksymets, Oleksandr and Zhao, Yili and Wijmans, Erik and Jain, Bhavana and Straub, Julian and Liu, Jia and Koltun, Vladlen and Malik, Jitendra and Parikh, Devi and Batra, Dhruv},
  title     = {{Habitat}: A Platform for Embodied {AI} Research},
  booktitle = {ICCV},
  year      = {2019},
  pages = {9339--9347}
}

@inproceedings{majumdar22zson,
  author    = {Majumdar, Arjun and Aggarwal, Gunjan and Devnani, Bhavika and Hoffman, Judy and Batra, Dhruv},
  booktitle = {NeurIPS},
  title     = {{ZSON}: Zero-Shot Object-Goal Navigation Using Multimodal Goal Embeddings},
  volume    = {35},
  year      = {2022},
  pages = {32340--32352}
}

@inproceedings{khanna2024goat,
  author    = {Khanna, Mukul and Ramrakhya, Ram and Chhablani, Gunjan and Yenamandra, Sriram and Gervet, Theophile and Chang, Matthew and Kira, Zsolt and Chaplot, Devendra Singh and Batra, Dhruv and Mottaghi, Roozbeh},
  title     = {{GOAT-Bench}: A Benchmark for Multi-Modal Lifelong Navigation},
  booktitle = {CVPR},
  year      = {2024},
  pages     = {16373--16383}
}

@inproceedings{yang2025embodiedbench,
  title     = {{EmbodiedBench}: Comprehensive Benchmarking Multi-modal Large Language Models for Vision-Driven Embodied Agents},
  author    = {Rui Yang and Hanyang Chen and Junyu Zhang and Mark Zhao and Cheng Qian and Kangrui Wang and Qineng Wang and Teja Venkat Koripella and Marziyeh Movahedi and Manling Li and Heng Ji and Huan Zhang and Tong Zhang},
  booktitle = {ICML},
  year = {2025},
  pages = {70576--70631},
  volume = {267}
}

@inproceedings{puig2024habitat3,
  title     = {{Habitat 3.0}: A Co-Habitat for Humans, Avatars, and Robots},
  author    = {Xavier Puig and Eric Undersander and Andrew Szot and Mikael Dallaire Cote and Tsung-Yen Yang and Ruslan Partsey and Ruta Desai and Alexander Clegg and Michal Hlavac and So Yeon Min and Vladim{\'\i}r Vondru{\v{s}} and Theophile Gervet and Vincent-Pierre Berges and John M Turner and Oleksandr Maksymets and Zsolt Kira and Mrinal Kalakrishnan and Jitendra Malik and Devendra Singh Chaplot and Unnat Jain and Dhruv Batra and Akshara Rai and Roozbeh Mottaghi},
  booktitle = {ICLR},
  year = {2024}
}

@inproceedings{shah2023vint,
  title     = {{ViNT}: A Foundation Model for Visual Navigation},
  author    = {Dhruv Shah and Ajay Sridhar and Nitish Dashora and Kyle Stachowicz and Kevin Black and Noriaki Hirose and Sergey Levine},
  booktitle = {CoRL},
  year = {2023},
  volume = {229},
  pages = {711--733}
}

@article{Zaffar2021VPRBench,
  author  = {Zaffar, Mubariz and Garg, Sourav and Milford, Michael and Kooij, Julian and Flynn, David and McDonald{-}Maier, Klaus and Ehsan, Shoaib},
  title   = {{VPR-Bench}: An Open-Source Visual Place Recognition Evaluation Framework with Quantifiable Viewpoint and Appearance Change},
  journal = {IJCV},
  year    = {2021},
  volume  = {129},
  pages = {2136--2174}
}

@inproceedings{zitkovich2023rt2,
  title = {{RT-2}: Vision-Language-Action Models Transfer Web Knowledge to Robotic Control},
  author = {Zitkovich, Brianna and Yu, Tianhe and Xu, Sichun and Xu, Peng and Xiao, Ted and Xia, Fei and Wu, Jialin and Wohlhart, Paul and Welker, Stefan and Wahid, Ayzaan and Vuong, Quan and Vanhoucke, Vincent and Tran, Huong and Soricut, Radu and Singh, Anikait and Singh, Jaspiar and Sermanet, Pierre and Sanketi, Pannag R. and Salazar, Grecia and Ryoo, Michael S. and Reymann, Krista and Rao, Kanishka and Pertsch, Karl and Mordatch, Igor and Michalewski, Henryk and Lu, Yao and Levine, Sergey and Lee, Lisa and Lee, Tsang-Wei Edward and Leal, Isabel and Kuang, Yuheng and Kalashnikov, Dmitry and Julian, Ryan and Joshi, Nikhil J. and Irpan, Alex and Ichter, Brian and Hsu, Jasmine and Herzog, Alexander and Hausman, Karol and Gopalakrishnan, Keerthana and Fu, Chuyuan and Florence, Pete and Finn, Chelsea and Dubey, Kumar Avinava and Driess, Danny and Ding, Tianli and Choromanski, Krzysztof Marcin and Chen, Xi and Chebotar, Yevgen and Carbajal, Justice and Brown, Noah and Brohan, Anthony and Arenas, Montserrat Gonzalez and Han, Kehang},
  booktitle = {CoRL},
  year      = {2023},
  volume    = {229},
  pages = {2165--2183}
}

@inproceedings{ichter2022do,
  title     = {Do As {I} Can, Not As {I} Say: Grounding Language in Robotic Affordances},
  author    = {Ichter, Brian and Brohan, Anthony and Chebotar, Yevgen and Finn, Chelsea and Hausman, Karol and Herzog, Alexander and Ho, Daniel and Ibarz, Julian and Irpan, Alex and Jang, Eric and Julian, Ryan and Kalashnikov, Dmitry and Levine, Sergey and Lu, Yao and Parada, Carolina and Rao, Kanishka and Sermanet, Pierre and Toshev, Alexander T and Vanhoucke, Vincent and Xia, Fei and Xiao, Ted and Xu, Peng and Yan, Mengyuan and Brown, Noah and Ahn, Michael and Cortes, Omar and Sievers, Nicolas and Tan, Clayton and Xu, Sichun and Reyes, Diego and Rettinghouse, Jarek and Quiambao, Jornell and Pastor, Peter and Luu, Linda and Lee, Kuang-Huei and Kuang, Yuheng and Jesmonth, Sally and Joshi, Nikhil J. and Jeffrey, Kyle and Ruano, Rosario Jauregui and Hsu, Jasmine and Gopalakrishnan, Keerthana and David, Byron and Zeng, Andy and Fu, Chuyuan Kelly},
  booktitle = {CoRL},
  pages = 	 {287--318},
  year = 	 {2023},
  volume = 	 {205}
}

@inproceedings{huang2022zeroshot,
  title = {Language Models as Zero-Shot Planners: Extracting Actionable Knowledge for Embodied Agents},
  author = {Huang, Wenlong and Abbeel, Pieter and Pathak, Deepak and Mordatch, Igor},
  booktitle = {ICML},
  pages = {9118--9147},
  year = {2022},
  volume = {162}
}

@inproceedings{li2024embodiedinterface,
  author    = {Li, Manling and Zhao, Shiyu and Wang, Qineng and Wang, Kangrui and Zhou, Yu and Srivastava, Sanjana and Gokmen, Cem and Lee, Tony and Li, Li Erran and Zhang, Ruohan and Liu, Weiyu and Liang, Percy and Fei-Fei, Li and Mao, Jiayuan and Wu, Jiajun},
  booktitle = {NeurIPS},
  title     = {{Embodied Agent Interface}: Benchmarking {LLMs} for Embodied Decision Making},
  volume    = {37},
  year      = {2024},
  pages = {100428--100534}
}

@article{kawaharazuka2025vla,
  author  = {Kawaharazuka, Kento and Oh, Jihoon and Yamada, Jun and Posner, Ingmar and Zhu, Yuke},
  journal = {IEEE Access},
  title   = {Vision-Language-Action Models for Robotics: A Review Towards Real-World Applications},
  year    = {2025},
  volume  = {13},
  pages = {162467--162504}
}

@inproceedings{Haas_2024_CVPR,
  author    = {Haas, Lukas and Skreta, Michal and Alberti, Silas and Finn, Chelsea},
  title     = {{PIGEON}: Predicting Image Geolocations},
  booktitle = {CVPR},
  year      = {2024},
   pages     = {12893--12902}
}

@inproceedings{dosovitskiy2017carla,
  title     = {{CARLA}: An Open Urban Driving Simulator},
  author    = {Dosovitskiy, Alexey and Ros, German and Codevilla, Felipe and Lopez, Antonio and Koltun, Vladlen},
  booktitle = {CoRL},
  year      = {2017},
  volume    = {78},
  pages = 	 {1--16},
}

@inproceedings{schumann-riezler-2021-generating,
  title     = {Generating Landmark Navigation Instructions from Maps as a Graph-to-Text Problem},
  author    = {Schumann, Raphael  and Riezler, Stefan},
  booktitle = {ACL-IJCNLP},
  year      = {2021},
  pages     = {489--502}
}

@inproceedings{
    wang2026cityseeker,
    title={{CitySeeker}: How Do {VLM}s Explore Embodied Urban Navigation with Implicit Human Needs?},
    author={Siqi Wang and Chao Liang and Yunfan Gao and Erxin Yu and Sen Li and Jing Li and Haofen Wang},
    booktitle={ICLR},
    year={2026},
}

@article{liu2026sidewalkbench,
  title   = {{SidewalkBench}: Benchmarking Visual Navigation on Urban Sidewalks},
  author={Zhizheng Liu and Honglin He and Vivek Alumootil and Akshat Pandya and Brad Squicciarini and Wayne Wu and Bolei Zhou},
  journal = {arXiv preprint arXiv:2606.16953},
  year    = {2026}
}

\appendix
\providecommand{\startsupplement}{\clearpage\appendix\onecolumn}
\startsupplement

In this supplement, we describe the experimental details in \Cref{sec:experiment_detail} and prompts used for the 360CityArena experiments in \Cref{sec:prompts}.

\section{Experiment Details}
\label{sec:experiment_detail}
\subsection{Task Settings}
For each task, a starting point and direction are defined. In addition, as described in Section~\ref{subsec:task_prompts}, prompts for each task type, as well as the variables and reference images for each task, are specified.
These tasks are designed to evaluate complementary aspects of embodied navigation, including localization, landmark grounding, instruction following, and spatial reasoning.

\subsection{Map Information}
The maps used in 360CityArena are created by cropping the exploration area from OpenStreetMap.

\subsection{Stopping Conditions}
In our experiments, we introduced the following stopping conditions. The details are as follows:

\begin{itemize}
    \item STEP\_LIMIT --- One action is counted as one step, and the process terminates when the number of steps exceeds 50.
    \item STAGNATION --- The process terminates when the agent repeats the same action 20 times consecutively without any change in position.
    \item AWAY\_FROM\_GOAL --- The process terminates when the agent moves away from the goal location for five consecutive movements, excluding viewpoint changes.
\end{itemize}

\subsection{Model Settings}
Claude Sonnet 4.5 and Qwen2.5-VL-32B-Instruct were evaluated with a temperature of 1.0, while GPT-5 does not allow temperature configuration. The maximum output length was set to 8{,}096 tokens.
In addition to the prompts described in Section~\ref{sec:prompts}, the model input consists of the viewpoint image, a map indicating the agent's current location, the Reflection Memory, and the task-specific reference image. The Reflection Memory is updated at every step and included in the input to the model. However, because Claude imposes a 5 MB input size limit, the current-location map is downscaled in 15\% steps as needed to stay under that limit.

\section{Prompts for 360CityArena Experiments}
\label{sec:prompts}
This section lists all the prompts used in our experiments. The prompts are divided into System Prompt, Reflection Prompt, and Task Prompt. In addition to these prompts, the inputs to the MLLMs include the viewpoint image, the map indicating the current location, the Reflection Memory, and the task-specific reference map.

\subsection{System Prompt}
The System Prompt specifies the reasoning procedure, the available actions, and the format of the output.
All prompts were designed specifically for this benchmark and were not part of the training data of the evaluated models.

\begin{tcblisting}{
  mylisting,
  title=System Prompt
}
You are a computer agent that uses the ReACT (Reasoning, Action, Observation) framework with memory to explore a city.
For each step, you should:
1. Think: Analyze the current state and decide what to do next
2. Action: Choose one of the following actions:
    - W: move forward
    - LEFT/RIGHT/UP: if you see red arrows, you should select one of them.
    - S: turn camera to the direction of travel
    - Q: rotate camera upward (look around)
    - E: rotate camera downward (look around)
    - A: rotate camera left (look around)
    - D: rotate camera right (look around)
    - ANSWER: answer the question

    NOTE:
    - If you see a big red arrow in front of you and want to go straight, you should select UP.
    - You cannot select LEFT, RIGHT or UP unless you see a big red arrow in front of you.
    - If pressing "W" does not move you forward, look around; red arrows will appear. You cannot move in any direction where a red arrow is not visible.

3. Observation: You will receive the result of your action

You will receive two types of images:
1. Camera view: The first-person view of what you can see in the city
2. Map view (when available): A top-down map showing your current location with a red arrow indicating your position and direction

Use both images to make better navigation decisions. The map can help you understand your location and plan your route more effectively.

Respond in the following JSON format:
{
    "thought": "your reasoning about what to do next",
    "action": "one of the available actions",
    "memory": "important information to remember for future steps",
    "answer": "the answer to the question of the task"
}

To not update memory, respond with an empty string.

For example:
{
    "thought": "I need to move forward",
    "action": "W",
    "memory": "1. My short term plan is to find the signboard of the road. 2. I need to move forward to find the signboard.",
    "answer": ""
}

Move control (W):
- Required: set "answer" to one of SMALL / MEDIUM / LARGE
- Mapping: SMALL = short move, MEDIUM = normal move, LARGE = long move
- Note: do not include anything else in "answer" when action is "W".

Rotation control (A/D/Q/E):
- Required: set "answer" to one of SMALL / MEDIUM / LARGE
- Mapping: SMALL ≈ 30°, MEDIUM ≈ 60°, LARGE ≈ 90°
- Default: if "answer" is omitted, ≈ 24° (≈ 0.5s at ~48°/s) is used.

Another example of changing direction:
{
    "thought": "I need to turn to the direction of travel",
    "action": "S",
    "memory": "",
    "answer": ""
}

Another example of answering the question:
{
    "thought": "",
    "action": "ANSWER",
    "memory": "The name of the city is Tokyo.",
    "answer": "x:100 y:100"
}


Do NOT wrap anything in ```json``` tags, and only respond with the JSON object.

Always analyze the screenshot carefully to determine the correct coordinates for your actions.
When a map is provided, use it to understand your current position and make more informed navigation decisions.
The memory field should contain any important information you want to remember for future steps.
\end{tcblisting}

\subsection{Reflection Prompt}
The Reflection Prompt is designed to provide a mechanism that enables the agent to maintain long-term consistency while planning and selecting actions. It specifies how information should be stored in the reflection memory. The reflection memory is saved as text and is updated after each action, serving as input to the agent in the next reasoning step.

\begin{tcblisting}{
  mylisting,
  title=Reflection Prompt
}
You will only see your last few observations and actions, so you will need to remember
important goals, objectives, and information that may be relevant. Make sure to read
all the text on the screen and use it to update your reflection memory!

You will be given a reflection memory that you can update with your current thoughts -- be careful NOT to overwrite your previous
reflection with a new one -- make sure to copy the previous reflection and add to it if you want to retain information. Do not be
conservative with your memory, you will need to remember everything!

Consider reflecting on:
- Important city objectives and goals
- Strategies that worked or didn't work
- Locations you've visited and what you found there
- Current status of the city

Think step by step and update your reflection memory with your current thoughts.
\end{tcblisting}

\subsection{Task Prompts}
\label{subsec:task_prompts}
The Task Prompts are prompts defined for each task type. For each task, certain variables and the input images are modified accordingly.

\paragraph{Localization}
\begin{figure}[t]
    \centering
    \includegraphics[width=0.9\linewidth]{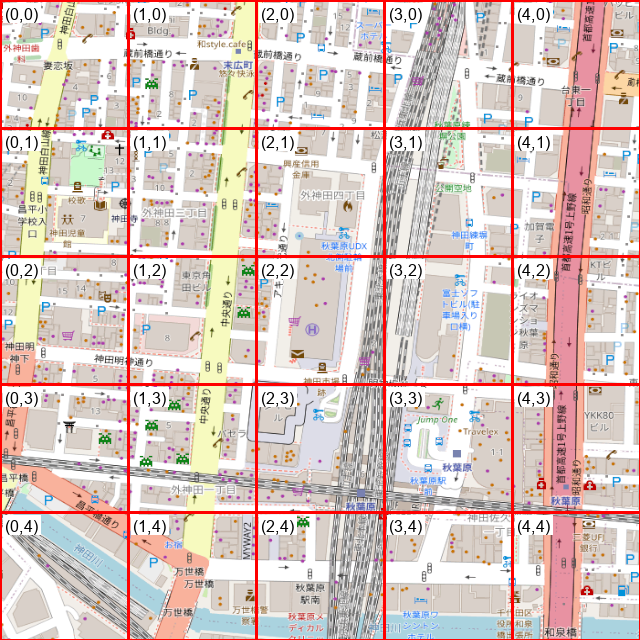}
    \caption{\textbf{Grid map provided in the Localization Task.} The task reference image in the Localization Task Prompt corresponds to this image. Map data \copyright\ OpenStreetMap contributors, ODbL 1.0, \url{https://www.openstreetmap.org/copyright}.}
    \label{fig:grid_map}
\end{figure}

In the Localization Task, the grid map shown in Figure~\ref{fig:grid_map} is provided, and the agent is required to estimate its current position from this map. The expected answer format is also specified within the prompt.

\begin{tcblisting}{
  mylisting,
  title=Localization Task Prompt
}
Your task is to explore the city and determine your initial starting position.
A reference map with a grid overlay is provided in the Task Reference Images. You MUST pick your answer from this grid: select the single grid cell that corresponds to your starting location and output its grid coordinate.

Before answering, actively explore your surroundings to gain confidence in your estimate: move around (walk a short distance), rotate your view (left/right and up/down), and re-check landmarks from multiple angles. Do not provide your answer until you are confident in your location.

You must complete your exploration and give your final answer within 50 steps.

When you have determined your starting position, specify "ANSWER" in the "action" field of your JSON response and provide your answer in the "answer" field in the format: "x:[grid_x] y:[grid_y]", where [grid_x] and [grid_y] are the integer indices of the selected grid cell from the provided grid. Do not output continuous coordinates (e.g., meters); only output the discrete grid indices.
\end{tcblisting}

\paragraph{Landmark Search with Language}
``LandmarkName'' is replaced with the name of a specific landmark for each task, such as ``Doutor Coffee'' or ``SEGA.'' To reduce task difficulty, we also include the information ``The goal is not far from the starting point.''

\begin{tcblisting}{
  mylisting,
  title=Landmark Search with Language Task Prompt
}
Your task is to go to {LandmarkName}. When you get in front of {LandmarkName}, use the ANSWER action to confirm completion.

The goal is not far from the starting point.
\end{tcblisting}

\paragraph{Landmark Search with Image}
\begin{figure}[t]
\centering
    \includegraphics[width=0.9\linewidth]{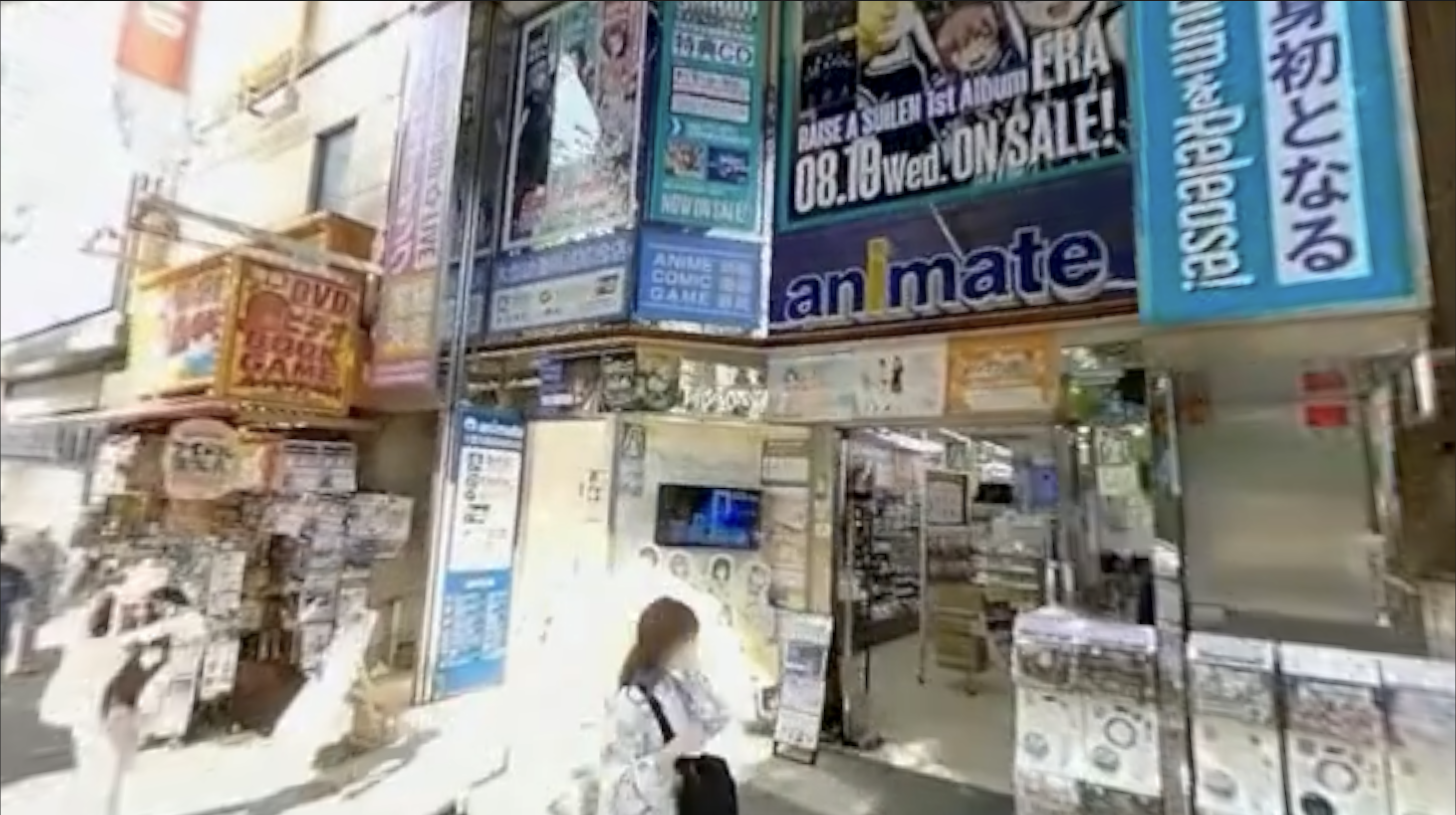}\\
   \caption{\textbf{Example of a landmark image provided in the Landmark  Search with Image task.} The task reference image in the Landmark Search with Image Task Prompt corresponds to this image.}
    \label{fig:landmark_search_example}
\end{figure}

In the Landmark Search with Image task, a task-specific landmark image, such as the one shown in Figure~\ref{fig:landmark_search_example}, is provided as the task reference image. The landmarks used here are the same ones selected in the ``Landmark Search with Language'' task.

\begin{tcblisting}{
  mylisting,
  title=Landmark Search with Image Task Prompt
}
Your task is to go to the landmark shown in the task reference image. When you get in front of the landmark, use the ANSWER action to confirm completion.

The goal is not far from the starting point.
\end{tcblisting}

\paragraph{Map Navigation}
\begin{figure}[t]
\centering
    \includegraphics[width=0.9\linewidth]{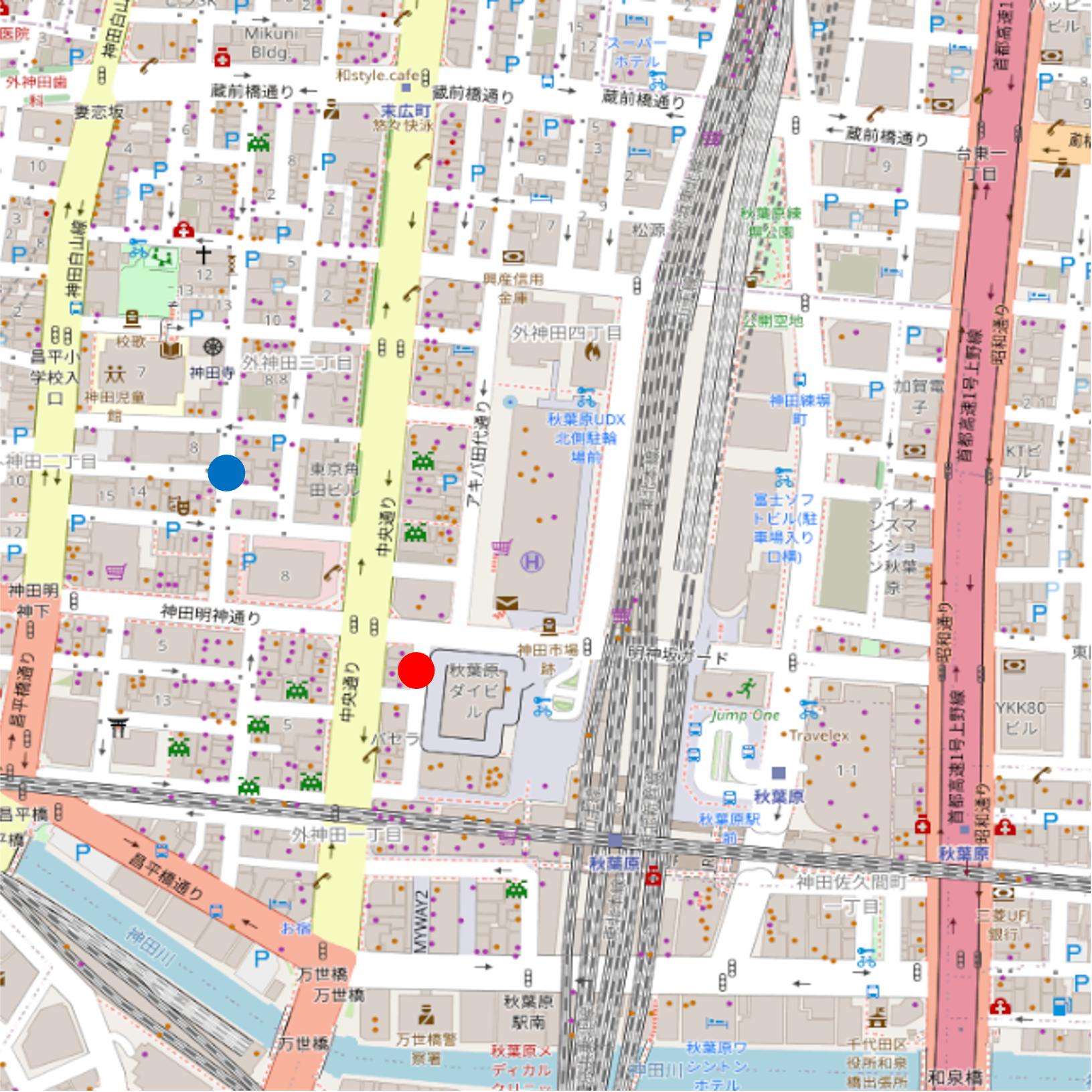}\\
    \caption{\textbf{Example of a map provided in the Map Navigation Task.} The task reference image in the Map Navigation Task Prompt corresponds to this image. Map data \copyright\ OpenStreetMap contributors, ODbL 1.0, \url{https://www.openstreetmap.org/copyright}.}
    \label{fig:map_nav_example}
\end{figure}

In the Map Navigation task, a task-specific map image such as the one shown in Figure~\ref{fig:map_nav_example} is provided as the reference image.

\begin{tcblisting}{
  mylisting,
  title=Map Navigation Task Prompt
}
Your task is to navigate from the starting position to the goal destination using the available actions.

A navigation reference map is provided in the task description above. On this reference map:
- BLUE marker indicates your starting position (initial location)
- RED marker indicates your goal destination

When you reach the goal area, use the ANSWER action to confirm completion.
\end{tcblisting}

\paragraph{Vision Language Navigation}
``Directions'' is replaced with task-specific navigation instructions such as the following.
\begin{enumerate}
\item Go straight.
\item Turn left at the first intersection.
\item Go straight.
\item Stop in front of Surugaya Purchase Center.
\end{enumerate}

\begin{tcblisting}{
  mylisting,
  title=Vision Language Navigation Task Prompt
}
Your task is to follow the directions to reach your destination. Please follow the instructions below:

{Directions}

Once you have reached your destination, output the ANSWER action.
\end{tcblisting}

\paragraph{Object Count}

``Object'' is replaced with a task-specific object name such as ``vending machine'' or ``planted tree.'' Likewise, ``Range'' is replaced with a description of the search area, such as ``the full perimeter of the block to your left'' or ``the road up to the next crosswalk.''

\begin{tcblisting}{
  mylisting,
  title=Object Count Task Prompt
}
Your task is to count the number of {Object} within {Range}.

Only count items that belong to the specified area/side/segment relative to your current position and along the described road segment. Do not include items outside the specified range.

If the specified range describes a block, count along the full perimeter of that block (all four sides) unless a specific side is explicitly specified (e.g., "right side only").

Output the answer as a number.
\end{tcblisting}

\paragraph{Relational Spatial Reasoning}

``LandmarkName'' is replaced with a task-specific landmark name such as ``Akiba no X,'' while ``Relation'' is replaced with expressions describing spatial relationships, such as ``the store to the left of it'' or ``the store across the street from it.''

\begin{tcblisting}{
  mylisting,
  title=Relational Spatial Reasoning Task Prompt
}
Find a nearby {LandmarkName} and tell me the name of {Relation}. {LandmarkName} is right nearby.
\end{tcblisting}

\subsection{Fuzzy Match Evaluation Prompt}
\label{subsec:fuzzy_match_prompt}
The \texttt{fuzzy\_match} metric (Section~{3.4} in the main paper) uses an LLM to judge whether the agent's answer is semantically equivalent to the ground truth. The following system prompt is given to the evaluator LLM (GPT-5), along with a user message containing both answers.

\begin{tcblisting}{
  mylisting,
  title=Fuzzy Match System Prompt
}
You are a helpful assistant.
You are given a user answer and an expected answer.
Please determine if the user answer is correct.

The answers do not have to match exactly word-for-word - If you determine that different names refer to the same landmark, return True.
\end{tcblisting}

\noindent The evaluator receives the following user message, where \texttt{\{user\_answer\}} and \texttt{\{expected\_answer\}} are replaced with the agent's output and the ground-truth answer, respectively.

\begin{tcblisting}{
  mylisting,
  title=Fuzzy Match User Message
}
User answer: {user_answer}
Expected answer: {expected_answer}

Output the answer in the following format: is_correct: True/False
\end{tcblisting}

\end{document}